\documentclass[lettersize,journal]{IEEEtran}

\makeatletter
\DeclareRobustCommand\onedot{\futurelet\@let@token\@onedot}
\def\@onedot{\ifx\@let@token.\else.\null\fi\xspace}

\def\eg{\emph{e.g}\onedot}

\def\etal{\emph{et al}\onedot}
\makeatother
\usepackage{xspace}

\usepackage{amsmath,amsfonts}
\usepackage{algorithmic}
\usepackage{array}
\usepackage[caption=false,font=normalsize,labelfont=sf,textfont=sf]{subfig}
\usepackage{textcomp}
\usepackage{stfloats}
\usepackage{url}
\usepackage{verbatim}
\usepackage{graphicx}
\usepackage{times}
\usepackage{epsfig}
\usepackage{graphicx}
\usepackage{amsmath}
\usepackage{amssymb}
\usepackage{pifont}
\usepackage{multirow}
\usepackage{booktabs}
\usepackage{xcolor,graphicx,float}%图片
\usepackage{pifont}
\usepackage{multirow}
\usepackage{booktabs}
\usepackage[numbers,sort&compress]{natbib}%解决如[2], [3]
\usepackage{amsmath}%公式
\usepackage{bm}
\usepackage{soul}

\def\BibTeX{{\rm B\kern-.05em{\sc i\kern-.025em b}\kern-.08em
    T\kern-.1667em\lower.7ex\hbox{E}\kern-.125emX}}
\usepackage{balance}
\begin{document}
\title{Learning Referring Video Object Segmentation from Weak Annotation}

\author{Wangbo~Zhao,
        Kepan~Nan, 
        Songyang~Zhang,
        Kai~Chen, 
        Dahua~Lin, 
        Yang~You 
  \IEEEcompsocitemizethanks{ 
  \IEEEcompsocthanksitem W. Zhao and Y. You are affiliated with the School of Computing, National University of Singapore, Singapore. Work done during W. Zhao’s internship at Shanghai AI Laboratory. E-mail: wangbo.zhao96@gmail.com, youy@comp.nus.edu.sg
  \IEEEcompsocthanksitem K. Nan is from the School of Automation, Northwestern Polytechnical University, Xi’an, China.  E-mail: nankpan@163.com
  \IEEEcompsocthanksitem S. Zhang, K. Chen, and D. Lin are with Shanghai AI Laboratory. 
  \IEEEcompsocthanksitem D. Lin is also with The Chinese University of Hong Kong, China. 
  \IEEEcompsocthanksitem W. Zhao and K. Nan contributed equally to this work.
  \IEEEcompsocthanksitem Y. You and K. Chen are corresponding authors. }
  }
  
\markboth{IEEE TRANSACTIONS ON IMAGE PROCESSING, ~Vol.~XXX, No.~XXX, XXX~XXX}%
{Shell \MakeLowercase{\textit{et al.}}: Bare Demo of IEEEtran.cls for IEEE Journals}

\maketitle

\begin{abstract} 
Referring video object segmentation (RVOS) is a task that aims to segment the target object in all video frames based on a sentence describing the object. Although existing RVOS methods have achieved significant performance, they depend on densely-annotated datasets, which are expensive and time-consuming to obtain. In this paper, we propose a new annotation scheme that reduces the annotation effort by $\times 8$ times, while providing sufficient supervision for RVOS. Our scheme only requires a mask for the frame where the object first appears and bounding boxes for the rest of the frames. Based on this scheme, we develop a novel RVOS method that exploits weak annotations effectively. Specifically, we build a simple but effective baseline model, SimRVOS, for RVOS with weak annotation. Then, we design a cross frame segmentation module, which uses the language-guided dynamic filters from one frame to segment the target object in other frames to thoroughly leverage the valuable mask annotation and bounding boxes. Finally, we develop a bi-level contrastive learning method to enhance the pixel-level discriminative representation of the model with weak annotation. We conduct extensive experiments to show that our method achieves comparable or even superior performance to fully-supervised methods, without requiring dense mask annotations.

% TIP修改完毕

% enable the model to learn pixel-level discriminative representation both from the mask annotation and pseudo masks.  
% Extensive experiments and ablative analyses demonstrate that our method is able to achieve competitive or even better performance without the demand for dense mask annotation.
% Our method consists of two main components: 1) A cross frame segmentation module that uses language-guided dynamic filters to segment the target object across different frames, exploiting both the mask and the bounding boxes. 2) A bi-level contrastive learning method that enhances the pixel-level discriminative representation of the model using the mask and pseudo masks. 

% Extensive experiments and ablative analyses show that our method is able to achieve competitive or even better performance without the demand for dense mask annotation.

%写到12行
% To relieve the burden of data annotation while maintaining sufficient supervision for segmentation, we propose a new annotation scheme, in which we label the frame where the object first appears with a mask and use bounding boxes for the subsequent frames.

% It achieves $8 \times$ times faster labeling speed than full annotation, promoting the application of RVOS in the real world. 

% Based on this scheme, we propose a method to effectively learn RVOS from weak annotations. 
\end{abstract}

% Note that keywords are not normally used for peerreview papers.

\section{Introduction} \label{sec:introduction}
\IEEEPARstart{R}{eferring} video object segmentation (RVOS), which aims to segment the target object in a video based on the natural language description, is an emerging visual understanding task. The ability to segment any text-referred objects makes RVOS different from the conventional unimodal video tasks like video object segmentation (VOS) \cite{perazzi2016benchmark, caelles2017one, xu2018youtube, oh2019video} or video instance segmentation (VIS) \cite{yang2019video, wu2022defense, yang2021crossover, qi2021occluded}, and be expected to have a broad range of real-world applications such as human-robot interaction\cite{qi2020reverie} and video editing\cite{li2020manigan, Wang_2022_CVPR}. 

RVOS has recently witnessed tremendous progress thanks to multi-modality interaction modeling \cite{botach2022end, wu2022language, zhao2022modeling} based on the Transformer \cite{vaswani2017attention} and deep convolutional networks \cite{he2016deep}. Most existing works\cite{botach2022end, wu2022language, ding2022language, wu2022multi, li2022you, zhao2022modeling} mainly focus on the fully-supervised RVOS setting, which requires pixel-level annotation and are typically data-hungry. Some recent attempts\cite{wu2022language, li2022you, liang2021rethinking} even utilize the image referring object segmentation datasets (\eg Ref-COCO \cite{yu2016modeling}) to alleviate the issue of insufficient data.  Building large-scale densely-annotated RVOS datasets is a possible solution, but it is costly and time-consuming, which impedes the development of RVOS.

\begin{figure}[!t]
    \centering
    \includegraphics[width=1\linewidth]{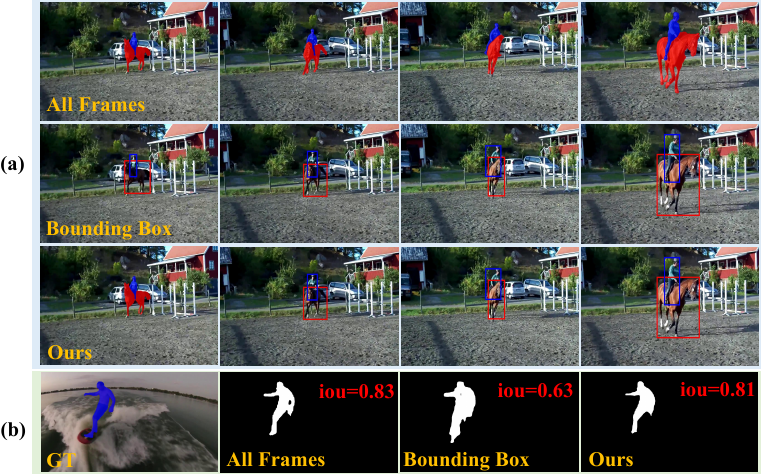}
    \caption{``All Frames'': All frames are labeled with masks. ``Bounding Box'': All frames are labeled with bounding boxes.  ``Ours'': We label the first frame where the target object first appears with a mask and other frames with bounding boxes. (a) Difference annotation schemes. (b) Predictions generated from a model trained with different annotation settings. ``Ours'' yields a more fine-grained mask than ``Bounding Box'' and is comparable with ``All Frames''. }
    \label{figure1}% √ % √ 
\end{figure}

In this work, we propose an alternative strategy to learn an RVOS model by leveraging the weak annotation. We first investigate the conventional weakly-supervised segmentation \cite{wang2022contrastmask, liu2021weakly, tian2021boxinst, song2019box} and find that the ``bounding boxes'' supervision not only extremely reduces the labeling cost but also is expected to enable the model to learn the robust and discriminative representation of the objects. However, due to the target of RVOS being to obtain accurate segmentation masks, using boxes as supervision for all frames may impose a non-negligible performance degradation. Thus, there exists a dilemma between reasonable mask prediction and annotation cost reduction in the RVOS task.

To cope with this dilemma, we introduce an improved labeling setting to adapt the RVOS. Specifically, we propose a new annotation scheme by labeling the segmentation mask for the frame where the object first appears and using bounding boxes for objects in the subsequent frames. The model learns to locate the target object from the bounding boxes and to produce fine-grained masks from the mask annotation during training. The newly introduced annotation scheme has two benefits: 
First, it enables state-of-the-art methods \eg ReferFormer \cite{wu2022language} to maintain competitive performance without the demand of dense annotation and is able to achieve $\times 8$ times faster labeling on datasets like YouTube-RVOS \cite{seo2020urvos} according to \cite{cheng2022pointly, papadopoulos2017extreme}. In addition, with high annotation efficiency, this annotation scheme can facilitate the construction of large-scale RVOS datasets in the future. Figure~\ref{figure1} compares different annotation schemes. The prediction mask from our scheme has a much higher IoU with the ground truth than that from ``Bounding Box”, and is close to that from ``All Frames”. We present more results and analysis in Section~\ref{Experiments}.

% The newly introduced annotation scheme enables state-of-the-art methods \eg ReferFormer \cite{wu2022language} to maintain competitive performance without the demand of dense annotation and is able to achieve $8\times$ times faster labeling on datasets like YouTube-RVOS \cite{seo2020urvos} according to \cite{cheng2022pointly, papadopoulos2017extreme}.

% In addition, this annotation scheme can facilitate constructing large-scale RVOS datasets with reduced labeling costs and preserved model performance. 

% locates the target object accurately and generates precise masks.

Based on the aforementioned annotation strategy, we then develop a novel and effective framework to learn from the weak annotation. Specifically, we first introduce a simple but effective model, SimRVOS. It adopts cross-modal fusion modules to prompt the interaction between visual and linguistic features. The prediction mask is generated by conducting a convolution operation on visual features with the dynamic filters generated from linguistic features. It can also serve as a baseline for future works for RVOS with weak annotation. Then, we design a language-guided cross frame segmentation (LGCFS) method, which uses the language-guided dynamic filters generated from other frames to segment the object in the current frame. This novel design not only enables us to use the valuable mask annotation to supervise the learning in all frames but also exploits the bounding box annotations as supervision. It also improves the robustness of the model against the appearance changes of the target object in the video. Finally, to encourage our model learning discriminative representation at the pixel level, we propose a bi-level contrastive learning (BLCL) method, which consists of (a) language-vision contrast and (b) consistency contrast, for model optimization. The former aligns the linguistic features with foreground visual features and separates them from background features, while the latter enhances the discrimination between foreground and background features. The BLCL accepts the supervision from both mask annotation and pseudo masks generated from bounding box annotation. We note that both LGCFS and BLCL can be easily applied to SimRVOS and other state-of-the-art RVOS methods, such as ReferFormer \cite{wu2022language}.

 % inspired by recent pixel-level contrastive learning works \cite{wang2021exploring, wang2022contrastmask, wang2022cris},

We conduct extensive experiments based on the proposed baseline SimRVOS and  ReferFormer\cite{wu2022language}. The empirical results and ablative analysis show our method is able to achieve superior performance on the representative benchmark, YouTube-RVOS \cite{seo2020urvos}, while significantly reducing the demand for expensive dense mask annotation. To summarize, the main contributions of our work are four-fold:
\begin{itemize}

    \item  We investigate different paradigms of weak supervision for RVOS and introduce a new weak annotation scheme.
    
    The newly proposed weak annotation scheme is promising to build an effective RVOS model with high annotation efficiency.

    \item We build a simple but effective model, SimRVOS, as the baseline for RVOS from weak annotation.

    \item We propose a language-guided cross frame segmentation method, to fully exploit the supervision from both mask and bounding boxes. And we design a bi-level contrastive learning paradigm to improve the discriminative representation at the pixel level with weak annotation.

    \item Experimental results demonstrate that our method is capable of improving the RVOS methods to achieve competitive or even better performance with only weak supervision. Finally, we indicate three potential directions for further work based on our weak annotation scheme.
\end{itemize}

% tip修改过introduction

\section{Related Work}
\begin{figure*}[!t]
  \centering
  \includegraphics[width=1\linewidth]{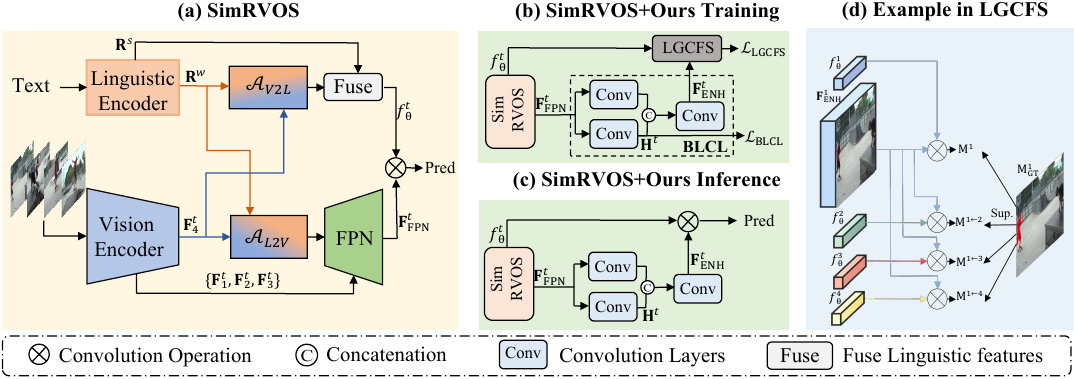}
%   \vspace{-2mm}
  \caption{The overview of the proposed model.``LGCFS'': Language-Guided Cross Frame Segmentation. ``BLCL'': Bi-Level Contrastive Learning. (a) The simple baseline SimRVOS. (b)(c) The pipeline of SimRVOS with proposed LGCFS and BLCL during training and inference. (d) For LGCFS, we illustrate the training process of the $1$-st frame in detail. Here, we assume that the $1$-st frame has a mask annotation. Language-guided dynamic filters from the $1$-st frame $f_{\theta}^1$ and other frames $f_{\theta}^2, f_{\theta}^3, f_{\theta}^4$ are applied to $\mathbf{F}_{\text{ENH}}^1$ to predict masks, so that the mask ground-truth $\mathbf{M}^1_{\text{GT}}$ from the $1$-st frame can supervise the learning in the $1$-st frame and other frames, simultaneously.}
  \label{network overview}   % √
\end{figure*}

\subsection{Referring Video Object Segmentation}  % pami版本修改完毕
Referring video object segmentation, also known as text-based video object segmentation, aims to find and segment the target object described by a text.  Gavrilyuk \etal \cite{gavrilyuk2018actor} first propose this task by extending two actor and action datasets \cite{xu2015can, jhuang2013towards} with natural language sentences. To increase the diversity in existing datasets, Seo \etal \cite{seo2020urvos} extend Youtube-VOS \cite{xu2018youtube} with language expressions describing the target object. To solve this task, some methods \cite{gavrilyuk2018actor, wang2020context} first extract linguistic features from the text and transform them into dynamic filters to convolve with vision features to generate segmentation masks. The other methods \cite{zhao2022modeling, ding2022language} try to introduce the motion information from the video into RVOS. Specifically, Zhao \etal \cite{zhao2022modeling} incorporate motion information from optical flow maps with appearance and linguistic features to segment the target object precisely. Ding \cite{ding2022language} adopt 2D ConvNet to extract features from frame difference to approximate motion information. With the success of vision transformers \cite{dosovitskiy2020image, liu2021swin}, recent efforts \cite{botach2022end,wu2022language} further improve the ROVS by either introducing the transformer structures. To be specific, Botach \etal \cite{botach2022end} processes frame and text features jointly in a multi-modal transformer, while Wu \etal \cite{wu2022language} views language as queries and adopt a transformer model to aggregate information from the most relevant regions in video frames. In addition to these methods, Liu \etal \cite{liu2021cross} extend the model for image referring object segmentation to a video version by processing a video frame by frame. Li \etal \cite{li2022you} build a meta-transfer module trained in a learning-to-learn manner to transfer the knowledge from the language domain to the image domain. Wu \etal \cite{wu2022multi} introduces multi-level alignment(video/frame/object-level) between the visual features and language features. Wu \etal \cite{wu2023onlinerefer} address RVOS in an online manner through explicit query propagation.

Although these methods have achieved significant performance, they rely heavily on densely-annotated datasets, which are costly to obtain. 
In contrast, our model can learn from weak annotations that are easier to obtain and achieve comparable or even better performance.

\subsection{Image/Video Segmentation with Weak Supervision} % pami版本修改完毕
Weakly-supervised methods for image semantic segmentation and instance segmentation have been widely studied, using various forms of supervision such as image label \cite{ahn2019weakly, jin2017webly, zhou2018weakly}, bounding box \cite{wang2022contrastmask, tian2021boxinst, song2019box, Dai_2015_ICCV, papandreou2015weakly}, point-level supervision \cite{laradji2020proposal, cheng2022pointly}, and scribble \cite{lin2016scribblesup, tang2018normalized}. Similarly, there are also many attempts to learn to segment pixels or objects in videos in a weakly-supervised manner \eg video semantic segmentation \cite{liu2021weclick, Hong_2017_CVPR, saleh2017bringing}, video instance segmentation \cite{liu2021weakly, ikeda2022weakly}, and video salient object detection \cite{zhao2021weakly, gao2022weakly}. However, for referring video object segmentation, the only existing weakly-supervised method is proposed by Chen \etal \cite{chen2022weakly}, who use an unsupervised tracker to obtain the target object’s bounding boxes and a pre-trained saliency segmentation model to generate pseudo labels. Their performance is significantly lower than that of fully-supervised methods due to the lack of fine-grained annotations.

Different from Chen \etal \cite{chen2022weakly}, our method achieves comparable or even better performance with state-of-the-art fully-supervised methods, with an acceptable burden of annotation.

\subsection{Video Object Segmentation}  % pami版本修改完毕
Conventionally, video object segmentation indicates the semi-supervised video object segmentation \cite{caelles2017one, perazzi2017learning, xu2018youtube, oh2019video}, where the first-frame mask is available during inference while \textit{densely-annotated frames are still needed in the training stage}. Hence, \emph{the ``semi-supervised'' here denotes the provided first-frame mask during inference rather than the learning manner.}  Therefore, some works \cite{Caelles_2017_CVPR, lu2020learning} also call this task one-shot VOS to avoid confusion.
The main challenge of VOS is how to exploit the first-frame mask effectively during inference. Existing methods can be classified into three categories: 1. Fine-tuning-based methods \cite{Caelles_2017_CVPR, voigtlaender2017online, bao2018cnn} adapt a pre-trained model to the target object using the first-frame mask before inference. However, these methods are computationally expensive and not suitable for real-time applications. 2. Propagation-based methods \cite{cheng2017segflow, luiten2018premvos, li2018video, chen2020state} propagate the mask in the first frame to the subsequent frames. 3. Matching-based methods \cite{oh2019video, voigtlaender2019feelvos, yang2020collaborative} store features from previous frames in a memory module and use them to segment the current frame by feature matching.

Different from video object segmentation, our method only needs the mask for the frame where the target object first appears and box annotations for other frames during training, which significantly reduces the burden of annotation. During inference, our model can segment the target object only based on the given text. \emph{We refer the reader to Section~\ref{annotation setting} for more comparisons between the RVOS with the proposed weak annotation scheme and VOS}.

\subsection{Video instance segmentation} % pami版本修改完毕
Video instance segmentation \cite{yang2019video}, is a challenging task that aims at simultaneously detecting, segmenting, and tracking object instances with pre-defined categories in videos. Methods for video instance segmentation can be categorized into online methods \cite{yang2019video, cao2020sipmask, yang2021crossover, qi2021occluded, liu2021sg, fang2021instances, wu2022defense, koner2023instanceformer} and offline methods \cite{wang2021end, seqformer, hwang2021video, heo2022vita, zhang2023towards}. Online methods detect and segment object instances for each frame and then perform frame-by-frame tracking. For example, Yang \etal \cite{yang2019video} attach a tracking branch to the model for image instance segmentation Mask R-CNN \cite{he2017mask}. Yang \etal \cite{yang2021crossover} conduct cross segmentation between frames during training to improve the robustness. Koner \etal \cite{koner2023instanceformer} propose a propagation mechanism and a memory module to model short-term and long-term dependencies in online VIS. These methods can process videos in real time but may suffer from object occlusions during object association. On the other hand, offline methods take a video clip as input and generate a sequence of instance masks directly. Specifically, VisTR \cite{wang2021end} inputs clip-level features into a transformer encoder \cite{carion2020end} and directly decoders a sequence of object masks in order. Heo \etal \cite{heo2022vita} propose an object decoder to associate frame-level predictions in a clip. Zhang \etal \cite{zhang2023towards} adopt the deformable attention \cite{zhu2020deformable} to aggregate spatial-temporal features from a video clip. However, these offline methods typically consume much more computational resources, especially for long videos.

In contrast to these methods for VIS, our method \textit{does not require pre-defined categories} and can perform class-agnostic segmentation with a referring expression.

\section{Our Approach}
In this work, we first introduce the formulation of learning RVOS from our weak annotation scheme in Section~\ref{Problem Formulation}. To learn a discriminative RVOS model with weak supervision, we then present a brief introduction of the baseline framework, termed SimRVOS, in Section~\ref{Framework}. It has a simple yet effective design, including a visual encoder, linguistic encoder, and cross-modal fusion modules. Based on the SimRVOS, we finally propose our two learning strategies to tackle the challenges in learning RVOS from weak annotations. The language-guided cross frame segmentation (LGCFS) and bi-level contrastive learning (BLCL) are detailed in Section~\ref{LGCFS} and \ref{CFCL}, respectively. We illustrate the overview of our SimRVOS in Figure~\ref{network overview}. We also visualize the pipeline of SimRVOS equipped with our LGCFS and BLCL during the training and inference stage.

\subsection{Problem Formulation} \label{Problem Formulation}
Referring video object segmentation aims at generating a series of segmentation masks $\mathcal{M}=\{\mathbf{M}^{t}\}_{t=1}^{T}$ for an object referred by a text $\mathcal{R}$ in a video clip  $\mathcal{I}=\{\mathbf{I}^{t}\}_{t=1}^{T}$, where $T$ is the number of frames in the video clip. Traditionally, we adopt a sequence of densely-annotated ground-truth masks $\mathcal{M}_{\text{GT}}=\{\mathbf{M}_{\text{GT}}^{t}\}_{t=1}^{T}$  to supervise the learning. However, pixel-level labeling is time-consuming and expensive to obtain. Hence, we propose to adopt a mask for the frame where the object first appears and bounding boxes in subsequent frames to supervise the model. Here, we assume that the target object appears in the first frame for convenience. Thus, our weakly-annotated ground-truth can be formulated as $\mathcal{W}_{\text{GT}}=\{\mathbf{M}_{\text{GT}}^{1}\} \cup \{\mathbf{B}_{\text{GT}}^{t}\}_{t=2}^{T}$, where $\mathbf{B}_{\text{GT}}^t$ represents the bounding box annotation of the target object in $t$-th frame.  % √ % √

\subsection{SimRVOS} \label{Framework}
We now give a brief introduction of the proposed simple baseline for RVOS, which consists of three modules:(a) visual encoder, (b) linguistic encoder, and (c) cross-modal fusion modules.  % √ % √

\noindent\textbf{Visual Encoder.}
Given a video clip, we input each frame $\mathbf{I}^{t}$ into a visual backbone and generate multi-scale appearance features $\{\mathbf{F}^t_i\}_{i=1}^{4}$, where $i$ represents the feature level. The superscript $t$ represents that the feature comes from the $t$-th frame. Both 2D backbones(\eg ResNet \cite{he2016deep} and Swin Transformer \cite{liu2021swin}) and 3D backbones(\eg Video Swin Transformer \cite{liu2022video}) can be used to generate features. % √% √

\noindent\textbf{Linguistic Encoder.}
Following \cite{wu2022language}, we input the referring text $\mathcal{R}$ into a language model to generate word-level features $\mathbf{R}^w=\{\mathbf{R}^w_{i}\}_{i=1}^{L}$ and sentence-level feature $\mathbf{R}^s$, where $L$ represents the length of the sentence. The sentence-level feature $\mathbf{R}^s$ can be viewed as a summarized representation of the whole sentence.  % √% √

\noindent\textbf{Cross-modal Fusion Modules.}
The fusion of appearance and linguistic features is essential to RVOS. To tackle this problem, we adopt two multi-head self-attention \cite{vaswani2017attention} modules to enable the interaction between the two modalities. One is language to vision attention $\mathcal{A}_{\text{L2V}}$, which is responsible for propagating linguistic features to vision features. The other is the vision to language attention $\mathcal{A}_{\text{V2L}}$, which is designed to enhance linguistic features with appearance information. They can be formulated as  
\begin{equation}
\hat{\mathbf{F}}^t=\mathcal{A}_{\text{L2V}}(\textbf{R}^t, \textbf{F}^t) = \textbf{F}^t + 
\mathcal{S}\left(\frac{f_q(\textbf{F}^t) f_k(\textbf{R}^t)^{\mathrm{T}}}{\sqrt{C}}\right) f_v(\textbf{R}^t),
\end{equation}
\begin{equation}
\hat{\mathbf{R}}^t = \mathcal{A}_{\text{V2L}}(\textbf{F}^t, \textbf{R}^t) = \textbf{R}^t + 
%\operatorname{softmax}
\mathcal{S}\left(\frac{f_q(\textbf{R}^t) f_k(\textbf{F}^t)^{\mathrm{T}}}{\sqrt{C}}\right) f_v(\textbf{F}^t),
\end{equation}
where $f_q$, $f_k$, and $f_v$ are linear functions to project original features to query, key, and value in the self-attention, respectively. $\mathcal{S}$ is the softmax activation operation. $\textbf{F}^t\in \mathbb{R}^{HW \times C}$ and $\textbf{R}^t \in \mathbb{R}^{L \times C}$are the visual and linguistic features input into the cross-modal fusion modules, respectively. In practice, we apply linear transformation both on the highest level visual feature $\mathbf{F}_4^t$ and word-level linguistic feature $\mathbf{R}^w$ to align their channels. We then view $\mathbf{F}_4^t$ and $\mathbf{R}^w$ as $\textbf{F}^t$ and $\textbf{R}^t$, respectively, and input them into the cross-modal fusion modules. This process is illustrated in Figure~\ref{network overview} (a). $\mathcal{A}_{\text{L2V}}$ and $\mathcal{A}_{\text{V2L}}$ share the similar structure but with individual parameters.

We then fuse the sentence-level feature $\mathbf{R}^s$ with the weighted average feature of the enhanced world-level feature $\hat{\mathbf{R}}^t$, to generate language-guided dynamic filters $f_{\theta}^t$ for each frame. For the vision branch, we generate the final feature representation $\mathbf{F}_{\text{FPN}}^t$ by adopting the feature pyramid network (FPN)  \cite{lin2017feature} to fuse the language-aware visual feature $\hat{\mathbf{F}}^t$ with multi-scale features from the vision encoder. Formally, we have:% √
\begin{equation}
f_{\theta}^t = \mathcal{G}_\text{MLP}(\mathbf{R}^s+\sum^L_{i=0}\lambda_i\hat{\mathbf{R}}^t_i),
\end{equation}
\begin{equation}
\mathbf{F}_{\text{FPN}}^t = \mathcal{G}_\text{FPN}(\hat{\mathbf{F}}^t, \mathbf{F}_1^t,\mathbf{F}_2^t,\mathbf{F}_3^t)
\end{equation}
where $\lambda_i$ is generated by passing $\hat{\mathbf{R}}^t$ through a linear function followed by a softmax function. $\mathcal{G}_{\text{MLP}}$ indicates a MLP block. 

In SimRVOS, we predict the final mask predictions by conducting the convolution on $\mathbf{F}_{\text{FPN}}^t$ with the generated filter $f_{\theta}^t$, which can be formulated as:
\begin{equation} \label{fpn seg eq}
\mathbf{M}^{t} = \mathcal{G}_\text{conv}(\mathbf{F}_{\text{FPN}}^{t}; f_{\theta}^{t}),
\end{equation}
where $\mathcal{G}_\text{conv}$ is the convolution operation. Figure~\ref{network overview} illustrates the process.

We will present the details of SimRVOS equipped with LGCFS and BLCL in the following two sections.  When our SimRVOS equips with LGCFS and BLCL, mask predictions are generated by the convolution operation between dynamic filters $f_{\theta}^t$ and contrastive learning enhanced feature $\mathbf{F}_{\text{ENH}}^t$, which will be detailed in Section.\ref{CFCL}. The processes of training and inference are illustrated in Figure~\ref{network overview} (b) and (c), respectively.

\subsection{Language-Guided Cross Frame Segmentation} \label{LGCFS}

% To leverage the only mask annotation in WS-RVOS, we introduce the cross frame segmentation strategy with the language-guided dynamic filters. 
We denote the predicted mask with langauge-guided filter as follows:
\begin{equation} \label{seg eq}
\mathbf{M}^{t} = \mathcal{G}_\text{conv}(\mathbf{F}_{\text{ENH}}^{t}; f_{\theta}^{t}).
% \vspace{-2mm}
\end{equation}
where $\mathcal{G}_\text{conv}$ is the convolution operation. $\mathbf{F}_{\text{ENH}}^t$ is the feature enhanced by our BLCL, and please refer to Section.\ref{CFCL} for more details. We typically use the ground-truth mask $\mathbf{M}_{\text{GT}}^{t}$ as the prediction target of $\mathbf{M}^t$ in the fully supervised learning RVOS. However, in the proposed weak annotation scheme, dense mask annotation is only available in the object's first appearing frame. Thus, we develop the cross segmentation method to maximize mining the only mask annotation. Specifically, we formulate the cross-frame segmentation as follows:% √% √
\begin{equation} 
\mathbf{M}^{t \leftarrow \tau} = \mathcal{G}_\text{conv}(\mathbf{F}_{\text{ENH}}^{t}; f_{\theta}^{\tau}), \tau \in [1, T] \text{ and } \tau \neq t,
\end{equation}
where $f_{\theta}^{\tau}$ represents the language-guided dynamic filters from the $\tau$-th frame. The cross segmentation enables the mask ground truth of the $t$-th frame to provide the supervision signal for other frames within one video clip. This paradigm also improves the robustness of the RVOS model against appearance change and occlusions.  % √% √
% ensuring we can fully use the valuable mask annotation. Another advantage of LGCFS is that it forces the learned language-guided dynamic filters to be robust to the appearance change of the object and object occlusions. 

For the frame with mask ground-truth, we adopt the combination of Dice loss \cite{milletari2016v} and Focal loss \cite{lin2017focal} as the mask loss $\mathcal{L}_{\text{MASK}}$ to optimize the model. It can be formulated as follows:
% Thus, we develop the cross segmentation method inspired by the \cite{yang2021crossover}
% \begin{equation} \label{seg loss}
% L_{\text{seg}}^{t} = L_{\text{mask}}(M^{t}, M^{t}_{\text{GT}}) + \sum_{\tau \in [1, T], \tau \neq t} L_{\text{mask}}(M^{t \leftarrow \tau}, M^{t}_{\text{GT}})
% \end{equation}
% \vspace{-3mm}
\begin{equation} \label{seg loss}
\mathcal{L}_{\text{LGCFS}}^{t} = \sum_{\tau \in [1, T], \tau \neq t} \mathcal{L}_{\text{MASK}}(\mathbf{M}^{t \leftarrow \tau}, \mathbf{M}^{t}_{\text{GT}})
\end{equation}
And the details of $\mathcal{L}_{\text{MASK}}$ are represented in Section~\ref{Loss Function in Our Method}.

However, in our weak annotation scheme, most of the frames only have the bounding box annotation.  To tackle this, we take the multiple instance learning (MIL) loss to constrain the predicted mask within the bounding box for these frames. The training data in MIL loss contains positive bags and negative bags. If a bag only contains negative instances, it is viewed as a negative bag, while is a positive bag. In our bounding box annotation, a vertical or a horizontal crossing line within the bounding box is viewed as a positive bag since it contains at least one foreground pixel. While other lines out of the bounding box are negative bags. Please find more details in \cite{NEURIPS2019_e6e71329, tian2021boxinst}. Based on this, the loss function in frames only with bounding box annotations can be formulated as follows:% √
\begin{equation} \label{MIL loss}
\mathcal{L}_{\text{LGCFS}}^{t} =  \sum_{\tau \in [1, T], \tau \neq t} \mathcal{L}_{\text{MIL}}(\mathbf{M}^{t \leftarrow \tau}, \mathbf{B}^{t}_{\text{GT}}).
\end{equation}% √% √
Here, the MIL loss constrains the predicted mask $\mathbf{M}^{t\leftarrow \tau}$ within the bounding box $\mathbf{B}^{t}_{\text{GT}}$.

%and is widely adopted in box-supervised tasks \cite{NEURIPS2019_e6e71329, tian2021boxinst}

According to Equations~\ref{seg loss} and~\ref{MIL loss}, the dynamic filters from all frames are constrained to predict masks not only for the mask-labeled frame but also for the box-labeled frames, avoiding overfitting to the mask of the mask-labeled frame.

We illustrate the training process of LGCFS in Figure~\ref{network overview} (d) and adopt the $1st$-th frame with the mask annotation as an example. The language-guided dynamic filters $f^1_{\theta}, f^2_{\theta}, f^3_{\theta}, f^4_{\theta}$ generated from different frames are used to predict the mask of the $1$-st frame. We note that the LGCFS is only conducted during training and we only adopt Equation~\ref{seg eq} to predict masks in the inference stage. Figure~\ref{network overview} (c) demonstrates this process.

\subsection{Bi-level Contrastive Learning} \label{CFCL}

\begin{figure}[!t]
    \centering
    \includegraphics[width=1\linewidth]{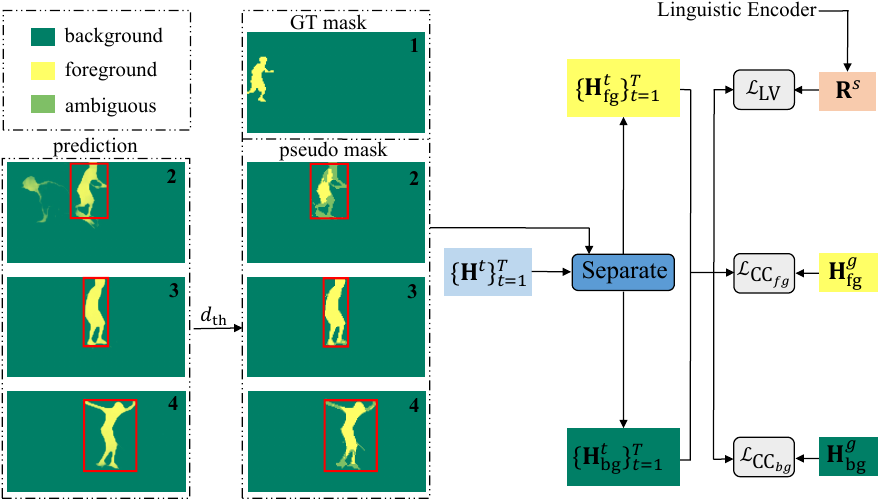}
    \caption{The pipeline of bi-level contrastive learning (BLCL). We adopt a threshold $d_{\text{th}}$ and bounding box annotations to convert predictions to pseudo masks for frames without a ground-truth mask.}
    \label{figure3} 
\end{figure}

We further develop a bi-level contrastive learning strategy to encourage our RVOS model to learn more discriminative representation under the weakly supervision. It consists of  (1) \textit{language-vision contrast}: the linguistic features should be close to the foreground vision features and far away from the background vision features in all frames, and (2) \textit{consistency contrast}: the global representation of the foreground region should be close to all foreground features and far away from background features in all frames, and vice versa.
The pipeline of our BLCL is shown in Figure.~\ref{figure3}.% √% √

% Inspired by some pixel-level contrastive learning methods \cite{wang2021exploring, wang2022contrastmask, wang2022cris},

\noindent\textbf{Separating foreground-background features.}
We first introduce the definition of foreground and background features in our BLCL. Specially, we transform the $\mathbf{F}_{\text{FPN}}^t$ into  $\mathbf{H}^t$ with several convolution layers for BLCL. For the frame with mask annotation, we use the downsampled mask to guide the separation between the foreground and background features, as follows: 
\begin{equation}
\mathbf{H}_{\text{fg}}^{t} = \left\{\mathbf{H}_{i,j}^{t} \mid \mathbf{M}_{\text{GT}_{i,j}}^{t}=1\right\}, 
\end{equation}
\begin{equation}
\mathbf{H}_{\text{bg}}^{t} = \left\{\mathbf{H}_{i,j}^{t} \mid \mathbf{M}_{\text{GT}_{i,j}}^{t}=0\right\},
\end{equation}
where the $(i,j)$ is the spatial coordinates, $\mathbf{H}_{\text{fg}}^t$ and $\mathbf{H}_{\text{bg}}^t$ are foreground and background features, respectively.
However, for those frames with only bounding box annotations, we are not able to separate foreground-background features precisely in a similar way. To solve this problem, we further introduce the prediction generated by Equation~\ref{seg eq} to generate the pseudo mask, to compensate for the box annotation softly. This progress can be formulated as:
\begin{equation}
\mathbf{H}_{\text{fg}}^{t} = \left\{\mathbf{H}_{i,j}^{t} \mid \mathcal{C}(\mathbf{M}_{{i,j}}^{t}, \mathbf{B}_{\text{GT}}^{t}) > d_\text{th} \right\},
\end{equation}
\begin{equation}
\mathbf{H}_{\text{bg}}^{t} = \left\{\mathbf{H}_{i,j}^{t} \mid \mathcal{C}(\mathbf{M}_{{i,j}}^{t}, \mathbf{B}_{\text{GT}}^{t}) < 1-d_\text{th} \right\}.
\end{equation}
where $\mathcal{C}(\mathbf{M}_{{i,j}}^{t}, \mathbf{B}_{\text{GT}}^{t})$ indicates the prediction score of the location $(i,j)$, which lies in the intersection of the pseudo mask and bounding box annotation. The locations whose score is larger than the pre-defined threshold $d_\text{th}$(\eg 0.9) will be considered as foreground locations. Location with a score lower than $1-d_\text{th}$ is viewed as a background location. Others are ambiguous regions and will be ignored in our BLCL. Apart from this, features locating out of box $\mathbf{B}_{\text{GT}}^{t}$ are viewed as background features directly and are appended into $\mathbf{H}_{\text{bg}}^{t}$ .

\noindent\textbf{Loss Definition.}
We now elaborate on the contrastive loss function used in our BLCL. It can be formulated as:
\begin{equation}
\begin{split}
\mathcal{L}\left(Q, K^{+} \mid K^{-} \right)  = \quad \quad \quad \quad \quad \quad \quad \quad  \quad \quad \quad \quad \\ 
\sum_{k\in K^{+} \cup K^{-}} \begin{cases}-\log \sigma (Q \cdot k), & k \in K^{+}, \\ -\log (1-\sigma (Q \cdot k)), & k \in K^{-} \end{cases}
\end{split}
\end{equation}
where $Q$ and $K^{+}$ are thought to be positive pairs, while $Q$ and $K^{-}$ are negative pairs. $\sigma$ denotes the sigmoid function. The loss should be averaged by the number of elements in $K^{+} \cup K^{-}$, which is omitted here. 
% This is similar to the definition in \cite{wang2022cris}.

For language-vision contrast, we adopt the linguistic feature $\mathbf{R}^{s}$ and foreground features $\{\mathbf{H}_{\text{fg}}^{t}\}_{t=1}^T$ to construct positive pairs, while $\mathbf{R}^{s}$ and background features $\{\mathbf{H}_{\text{bg}}^{t}\}_{t=1}^T$ are viewed as negative pairs. Thus, the language-vision contrast loss can be defined as: 
\begin{equation}
\mathcal{L}_{\text{LV}} = \mathcal{L}\left(\mathbf{R}^{s}, \{\mathbf{H}_{\text{fg}}^{t}\}_{t=1}^T \mid \{\mathbf{H}_{\text{bg}}^{t}\}_{t=1}^T \right).
\end{equation}

For consistency contrast, we take the averaged foreground/background features over all frames $\mathbf{H}_{\text{fg}}^{g}$ and $\mathbf{H}_{\text{bg}}^{g}$ as the anchors, which can be viewed as the global representation of foreground/background features. The global foreground feature and all frames' foreground features should be positive pairs, while it is negative with all frames' background features. The global background feature also follows this rule to build positive and negative pairs. We formulate the consistency contrast loss as $\mathcal{L}_{\text{CC}_{fg}}$ and $\mathcal{L}_{\text{CC}_{bg}}$:
\begin{equation}
\mathcal{L}_{\text{CC}_{fg}} = \mathcal{L}\left(\mathbf{H}_{\text{fg}}^{g}, \{\mathbf{H}_{\text{fg}}^{t}\}_{t=1}^T \mid \{\mathbf{H}_{\text{bg}}^{t}\}_{t=1}^T  \right)
\end{equation}

\begin{equation}
\mathcal{L}_{\text{CC}_{bg}} = \mathcal{L}\left(\mathbf{H}_{\text{bg}}^{g}, \{\mathbf{H}_{\text{bg}}^{t}\}_{t=1}^T \mid \{\mathbf{H}_{\text{fg}}^{t}\}_{t=1}^T  \right).
\end{equation}

% For language-vision contrast, we adopt the linguistic feature $\mathbf{R}^{s}$ and foreground features $\{\mathbf{H}_{\text{fg}}^{t}\}_{t=1}^T$ to construct positive pairs, while $\mathbf{R}^{s}$ and background features $\{\mathbf{H}_{\text{bg}}^{t}\}_{t=1}^T$ are viewed as negative pairs. We denote the language-vision contrast loss as $\mathcal{L}_{\text{LV}}$.

% For consistency contrast, we take the averaged foreground/background features over all frames $\mathbf{H}_{\text{fg}}^{g}$ and $\mathbf{H}_{\text{bg}}^{g}$ as the anchors, which can be viewed as the global representation of foreground/background features. 
% The global foreground feature and all frames' foreground features should be positive pairs, while it is negative with all frames' background features. The global background feature also follows this rule to build positive and negative pairs. We formulate the consistency contrast loss as $\mathcal{L}_{\text{CC}_{fg}}$ and $\mathcal{L}_{\text{CC}_{bg}}$ for global foreground and global background, respectively.  

Finally, the total BLCL loss function can be defined as: 
\begin{equation}
\mathcal{L}_{\text{BLCL}} = \mathcal{L}_{\text{LV}} + \mathcal{L}_{\text{CC}_{fg}} + \mathcal{L}_{\text{CC}_{bg}} 
\end{equation}
By introducing our BLCL, both frames with mask and bounding box annotations can learn foreground-background discrimination in a unified manner and contribute to the optimization of the segmentation.

Finally, we pass the original $\mathbf{F}_{\text{FPN}}^t$ feature through a convolutional layer and concatenate it with the feature $\mathbf{H}^t$, which has well learned foreground-background discrimination ability, followed by another two convolutional layers to reduce the channel dimension, generating the feature $\mathbf{F}_{\text{ENH}}^t$. These can be found in Figure~\ref{network overview} (b) and (c).

\section{Total Loss Function} \label{Loss Function in Our Method} 
In this section, We illustrate the total loss function in our method during training in detail. 

We adopt the combination of Dice loss \cite{milletari2016v} and Focal loss \cite{lin2017focal} to supervise the learning in Frame $t$, when it has the mask annotation. To balance the training, we set the weight of dice loss and focal loss to 5 and 2, respectively.
\begin{equation}
\begin{split}
\mathcal{L}_{\text{MASK}}(\mathbf{M}^{t}, \mathbf{M}^{t}_{\text{GT}}) = 5\mathcal{L}_{\text{Dice}}(\mathbf{M}^{t}, \mathbf{M}^{t}_{\text{GT}}) +  2\mathcal{L}_{\text{Focal}}(\mathbf{M}^{t}, \mathbf{M}^{t}_{\text{GT}}),
\end{split}
\end{equation}

When it comes to frames only with bounding boxes, we use the multiple instance learning loss, which is denoted as $\mathcal{L}_{\text{MIL}}$.  Hence, the common segmentation loss can be defined as:
\begin{equation}
\mathcal{L}_{\text{SEG}}^{t} = \begin{cases} \mathcal{L}_{\text{MASK}}(\mathbf{M}^{t}, \mathbf{M}^{t}_{\text{GT}}), & \text{w/ mask}, \\
\mathcal{L}_{\text{MIL}}(\mathbf{M}^{t}, \mathbf{B}^{t}_{\text{GT}}), & \text{w/o mask}, \end{cases}
\end{equation}
% \cite{NEURIPS2019_e6e71329, tian2021boxinst}
In Section~\ref{LGCFS}, we have introduced the loss function in LGCFS to leverage the annotation in Frame $t$ to supervise the learning process in other frames, denoted as $\mathcal{L}_{\text{LGCFS}}^t$. The weights of loss functions in our LGCFS are consistent with that in $\mathcal{L}_{\text{SEG}}^{t}$. The loss function in our BLCL is defined on the whole video clip, denoted as $\mathcal{L}_{\text{BLCL}}$, which has been described in Section~\ref{CFCL}. Hence, the total loss function can be defined as:
\begin{equation}
\mathcal{L}_{\text{Final}} = \sum_{t=1}^{T} \left (\mathcal{L}_{\text{SEG}}^{t} + \mathcal{L}_{\text{LGCFS}}^t\right) + \mathcal{L}_{\text{BLCL}}.
\end{equation}

\section{Experiments}\label{Experiments}
\subsection{Experimental Configuration}
\noindent\textbf{Datasets.}
For the dataset, we build our weakly-annotated dataset based on YouTube-RVOS \cite{seo2020urvos}. Concretely, we keep the mask and the bounding box annotation in the frame where the target object first appears and remove masks in other frames, resulting in only bounding box annotations in other frames. An example is shown in Figure~\ref{figure1} (a) ``Ours''. We use the same training and validation set as YouTube-RVOS \cite{seo2020urvos}, which contains 3,471 and 202 videos, respectively. Following \cite{wu2022defense, liu2021weakly}, we further hold out a subset of videos from the training split by randomly sampling 200 videos, resulting in $\text{train\_train}$ and $\text{train\_val}$ splits to conduct comprehensive evaluation and analysis on our method. For the convenience of comparison and reference, we will release the splits with our code. We also evaluate our method on DAVIS17-RVOS \cite{khoreva2018video}, which contains 30 videos for validation.

\begin{table}[t!]
	\centering
    \footnotesize

\caption{Comparison with state-of-the-art methods on the official validation set of YouTube-RVOS \cite{seo2020urvos}. ``*'' indicates that the second stage inference step is removed. $\dagger$ indicates that joint training on RefCOCO \cite{yu2016modeling} and YouTube-RVOS \cite{seo2020urvos} is adopted. }
  \vspace{-3mm}
\setlength{\tabcolsep}{3.3mm}{
	\begin{tabular}{lc|c|c|ccc|} 
	   %\hline
	    \toprule 
	          \multicolumn{2}{l|}{\multirow{2}{*}{\textbf{Method}}} & \multicolumn{1}{c|}{\multirow{2}{*}{\textbf{Super.}}}  &  \multicolumn{3}{c}{YouTube-RVOS val}  \\ 

	               \multicolumn{2}{l|}{} & \multicolumn{1}{c|}{}   & \multicolumn{1}{c}{$\mathcal{J} \& \mathcal{F}$} & \multicolumn{1}{c}{$\mathcal{J}$}  & \multicolumn{1}{c}{$\mathcal{F}$} \\  
	               \midrule

      	       	\multicolumn{2}{l|}{URVOS \cite{seo2020urvos}}   & \multicolumn{1}{c|}{Full}   & \multicolumn{1}{c}{47.2} & \multicolumn{1}{c}{45.3}      & \multicolumn{1}{c}{49.2}    \\ 
      	       	\multicolumn{2}{l|}{URVOS*  \cite{seo2020urvos}}   & \multicolumn{1}{c|}{Full}     & \multicolumn{1}{c}{-} & \multicolumn{1}{c}{41.3}      & \multicolumn{1}{c}{-}    \\ 
      	       	
     	       	\multicolumn{2}{l|}{CMPC-V \cite{liu2021cross}}    & \multicolumn{1}{c|}{Full}   & \multicolumn{1}{c}{47.5} & \multicolumn{1}{c}{45.6}      & \multicolumn{1}{c}{49.3}    \\

      	\multicolumn{2}{l|}{MLRL \cite{wu2022multi}}    & \multicolumn{1}{c|}{Full}        & \multicolumn{1}{c}{49.7} & \multicolumn{1}{c}{48.4}      & \multicolumn{1}{c}{50.9}    \\

      \multicolumn{2}{l|}{LBDT \cite{ding2022language}}   & \multicolumn{1}{c|}{Full}     & \multicolumn{1}{c}{49.4} & \multicolumn{1}{c}{48.2}      & \multicolumn{1}{c}{50.6}    \\ 
     	
     	\multicolumn{2}{l|}{YOFO$\dagger$ \cite{li2022you}}    & \multicolumn{1}{c|}{Full}    & \multicolumn{1}{c}{48.6} & \multicolumn{1}{c}{47.5}      & \multicolumn{1}{c}{49.7}    \\           \midrule

         	\multicolumn{2}{l|}{SimRVOS}      & \multicolumn{1}{c|}{Full}   & \multicolumn{1}{c}{44.7}    & \multicolumn{1}{c}{43.7}      & \multicolumn{1}{c}{45.8}    \\

            \multicolumn{2}{l|}{SimRVOS}      & \multicolumn{1}{c|}{$\text{Weak}_{\text{B}}$}   & \multicolumn{1}{c}{37.8}    & \multicolumn{1}{c}{38.6}      & \multicolumn{1}{c}{36.9}    \\

            \multicolumn{2}{l|}{SimRVOS}      & \multicolumn{1}{c|}{$\text{Weak}_{\text{M}}$}  & \multicolumn{1}{c}{40.9}    & \multicolumn{1}{c}{40.1}      & \multicolumn{1}{c}{41.8}    \\   
     	
      	\multicolumn{2}{l|}{SimRVOS}      & \multicolumn{1}{c|}{$\text{Weak}_{\text{M+B}}$}  & \multicolumn{1}{c}{42.7}    & \multicolumn{1}{c}{41.7}      & \multicolumn{1}{c}{43.6}    \\    
      	
      	\multicolumn{2}{l|}{SimRVOS+Ours}      & \multicolumn{1}{c|}{$\text{Weak}_{\text{M+B}}$}   & \multicolumn{1}{c}{46.6}    & \multicolumn{1}{c}{45.6}      & \multicolumn{1}{c}{47.6}    \\  \midrule

      	        \multicolumn{2}{l|}{ReferFormer \cite{wu2022language}}    & \multicolumn{1}{c|}{Full}   & \multicolumn{1}{c}{49.9} & \multicolumn{1}{c}{49.2}      & \multicolumn{1}{c}{50.6}    \\ 

      \multicolumn{2}{l|}{ReferFormer \cite{wu2022language}}    & \multicolumn{1}{c|}{$\text{Weak}_{\text{M+B}}$}    & \multicolumn{1}{c}{48.9} & \multicolumn{1}{c}{47.9}      & \multicolumn{1}{c}{49.9}    \\

        \multicolumn{2}{l|}{ReferFormer+Ours}      & \multicolumn{1}{c|}{$\text{Weak}_{\text{M+B}}$} & \multicolumn{1}{c}{50.1}    & \multicolumn{1}{c}{48.9}      & \multicolumn{1}{c}{51.2}    \\

      	\bottomrule
	\end{tabular}}

	\label{sota}
	% \vspace{-6mm}
\end{table}

\begin{table*}[h]
\centering
\caption{``VOS'': Semi-supervised video object segmentation. ``RVOS'': Fully-supervised referring video object Segmentation. ``WS-RVOS'': The proposed weak annotation scheme for referring video object Segmentation. ``text'' is the referring text in RVOS. ``Label'': The annotation for the target object. ``Guidance'': The guidance used to indicate the target object. VOS adopts the mask of the target object when it first appears, while RVOS and WS-RVOS only use the referring text.}
  \vspace{-3mm}
\setlength{\tabcolsep}{12.5mm}{\begin{tabular}{cccc}
\toprule 
Task & Label & Guidance & Labeling  Speed \\
             \midrule
VOS & $\text{mask}_{\text{all frames}}$  &  $\text{mask}_{\text{1st-frame}}$ &  $\times1$ \\
RVOS & $\text{mask}_{\text{all frames}}$   & text  &  $\times1$ \\
WS-RVOS & $\text{mask}_{\text{1st-frame}}$ + $\text{box}_{\text{other frames}}$   &  text &  $\times8$ \\
\bottomrule 
\end{tabular}}
\label{vos}
\vspace{-3mm}
\end{table*}

\begin{table*}[t!]
	\centering

    \footnotesize
	\renewcommand{\arraystretch}{1.}
	\renewcommand{\tabcolsep}{1.5mm}
 
\caption{Ablation studies on the main components of the proposed method. We evaluate models on official validation sets of YouTube-RVOS \cite{seo2020urvos} and DAVIS17-RVOS \cite{khoreva2018video}. \textbf{Blod} indicates the best score.}
   \vspace{-3mm}

	\begin{tabular}{cc|ccc|ccc} 

	\toprule 
	   \multicolumn{2}{c|}{\multirow{3}{*}{Name}}  &  \multicolumn{3}{c|}{YouTube-RVOS val}   &  \multicolumn{3}{c}{DAVIS17-RVOS}  \\
    
           \multicolumn{2}{c|}{}   & \multicolumn{1}{c}{$\mathcal{J} \& \mathcal{F}$}   & \multicolumn{1}{c}{$\mathcal{J}$} & \multicolumn{1}{c|}{$\mathcal{F}$} & \multicolumn{1}{c}{$\mathcal{J} \& \mathcal{F}$}   & \multicolumn{1}{c}{$\mathcal{J}$} & \multicolumn{1}{c}{$\mathcal{F}$} \\  \midrule
   
        \multicolumn{2}{c|}{SimRVOS}      & \multicolumn{1}{c}{42.7}  & \multicolumn{1}{c}{41.7} & \multicolumn{1}{c|}{43.6}   & \multicolumn{1}{c}{43.5}    & \multicolumn{1}{c}{40.8}   & \multicolumn{1}{c}{46.1} \\
        
        \multicolumn{2}{c|}{SimRVOS+LGCFS}       & \multicolumn{1}{c}{45.7}  & \multicolumn{1}{c}{44.7} & \multicolumn{1}{c|}{46.7}    & \multicolumn{1}{c}{46.3}     & \multicolumn{1}{c}{43.5}   & \multicolumn{1}{c}{49.0}  \\ 
   
        \multicolumn{2}{c|}{SimRVOS+LGCFS+BLCL (SimRVOS+Ours)}    & \multicolumn{1}{c}{\textbf{46.6}}   & \multicolumn{1}{c}{\textbf{45.6}} & \multicolumn{1}{c|}{\textbf{47.6}}     & \multicolumn{1}{c}{\textbf{47.3}}   & \multicolumn{1}{c}{\textbf{44.6}}   & \multicolumn{1}{c}{\textbf{50.0}} \\ 
 	       		       	  	       
	    \bottomrule
     
	\end{tabular}
	\label{ablation_table_official}
 \vspace{-3mm}
\end{table*}

\begin{table*}[t!]
	\centering

    \footnotesize
	\renewcommand{\arraystretch}{1.}
	\renewcommand{\tabcolsep}{1.5mm}
 
	\caption{Ablation studies on main components in the proposed method. We evaluate models on the splitted train\_val set.}
  \vspace{-3mm}
	\begin{tabular}{cc|ccc|cccccc|cc} 

	\toprule 
	   \multicolumn{2}{c|}{\multirow{3}{*}{Name}}   &  \multicolumn{11}{c}{YouTube-RVOS train\_val}   \\   \cmidrule(lr){3-13} 
	   
	    \multicolumn{2}{c|}{}   &  \multicolumn{3}{c|}{Video Metrics} & \multicolumn{5}{c|}{Precision}  & \multicolumn{1}{c|}{mAP}   & \multicolumn{2}{c}{IoU} \\   
	                  
       \multicolumn{2}{c|}{}   & \multicolumn{1}{c}{$\mathcal{J} \& \mathcal{F}$} & \multicolumn{1}{c}{$\mathcal{J}$} & \multicolumn{1}{c|}{$\mathcal{F}$} & \multicolumn{1}{c}{P@0.5}               & \multicolumn{1}{c}{P@0.6} &\multicolumn{1}{c}{P@0.7}      & \multicolumn{1}{c}{P@0.8}  &\multicolumn{1}{c|}{P@0.9}  & \multicolumn{1}{c|}{0.5:0.95}   & \multicolumn{1}{c}{Over All} & \multicolumn{1}{c}{Mean} \\  \midrule

	   \multicolumn{2}{c|}{SimRVOS}     & \multicolumn{1}{c}{61.8}  & \multicolumn{1}{c}{62.9}  & \multicolumn{1}{c|}{60.7}  & \multicolumn{1}{c}{49.8}   & \multicolumn{1}{c}{41.9}   & \multicolumn{1}{c}{33.4}   & \multicolumn{1}{c}{22.7} & \multicolumn{1}{c}{8.6}  & \multicolumn{1}{|c|}{28.6}   & \multicolumn{1}{c}{51.5}    & \multicolumn{1}{c}{46.6}  \\

   	\multicolumn{2}{c|}{SimRVOS+LGCFS}      & \multicolumn{1}{c}{63.9}   & \multicolumn{1}{c}{64.7}   & \multicolumn{1}{c|}{63.2} & \multicolumn{1}{c}{56.0} & \multicolumn{1}{c}{49.2} & \multicolumn{1}{c}{40.8} & \multicolumn{1}{c}{28.5} & \multicolumn{1}{c}{11.7} & \multicolumn{1}{|c|}{34.3}   & \multicolumn{1}{c}{56.3}    & \multicolumn{1}{c}{50.7}  \\

  	\multicolumn{2}{c|}{SimRVOS+LGCFS+BLCL (SimRVOS+Ours)}    &  \multicolumn{1}{c}{\textbf{65.1}}   & \multicolumn{1}{c}{\textbf{66.0}} & \multicolumn{1}{c|}{\textbf{64.2}} & \multicolumn{1}{c}{\textbf{57.4}} & \multicolumn{1}{c}{\textbf{51.2}} & \multicolumn{1}{c}{\textbf{42.9}} & \multicolumn{1}{c}{\textbf{30.5}} & \multicolumn{1}{c}{\textbf{12.7}} & \multicolumn{1}{|c|}{\textbf{35.8}}  & \multicolumn{1}{c}{\textbf{56.7}}    & \multicolumn{1}{c}{\textbf{51.6}}  \\

	       	\bottomrule
	\end{tabular}
	\label{ablation_table_splitted}
 \vspace{-3mm}
\end{table*}

\noindent\textbf{Evaluation Metrics.}
We adopt the region similarity $\mathcal{J}$, contour similarity $\mathcal{F}$ and their average value
$\mathcal{J} \& \mathcal{F}$ provided by \cite{seo2020urvos} to evaluate our method. Region similarity $\mathcal{J}$ measures the mean IoU between predictions and ground truth. $\mathcal{F}$ evaluates the similarity from a contour-based perspective. Both of these metrics evaluate results at the video level. Since the ground truth in $\text{train\_val}$ is available, we also evaluate our method with frame-level metrics: Precision $P@X$ and mean average precision (mAP) in this split for comprehensive analysis. $P@X$ measures the percentage of predicted masks whose IoU with ground-truth are higher than the threshold $X$, where $X$ $\in [0.5, 0.6, 0.7, 0.8, 0.9]$. mAP is the average of Precision $P@[0.5:0.95:0.05]$.

\noindent\textbf{Implementation Details.}
We adopt ResNet-50 \cite{he2016deep} as the visual encoder. Following previous methods \cite{wu2022language, botach2022end}, we assume RoBERTa \cite{liu2019roberta} with frozen parameters during training as our linguistic encoder. We select AdamW \cite{loshchilov2017decoupled} as our optimizer. The learning rate of the visual backbone is set to  2.5e-5, and the rests are set to 5e-5. We sample 8 video clips in each iteration, and each clip contains five frames. All experiments are conducted on 4 NVIDIA Tesla A100 GPUs. Unless otherwise specified, we train our model by 18 epochs and the learning rate is decayed at the $9$-th and $15$-th epochs. During training, we set the maximum size of the long side to 640, and the size of the short side is randomly selected from [288, 320, 352, 392, 416, 448, 480, 512]. We adopt random horizontal flips, random resize, random crop, and photometric distortion to augment the data. During inference, frames are downsampled to 360p. These training and inference settings are consistent with \cite{wu2022language}.  We introduce the pseudo mask in BLCL after finishing the first epoch training so that we can obtain high-quality pseudo masks. The threshold $d_{\mathrm{th}}$ in BLCL is set to 0.9 by default.

% In this section, we will briefly introduce the experiment configurations and more details(datasets, metrics, model and training recipes) are presentaed in Appendix.
% For the dataset, we build our weakly-annotated dataset based on YouTube-RVOS \cite{seo2020urvos}. Concretely, we keep the mask and the bounding box in the frame where the target object first appears and remove masks in other frames, resulting in only bounding box annotations in other frames. An example is shown in Figure.~\ref{figure1} (a) ``Ours". We use the same training and validation set as YouTube-RVOS \cite{seo2020urvos}. Following \cite{wu2022defense, liu2021weakly}, we further hold out a subset of videos from the training split by randomly sampling 200 videos, resulting $\text{train\_train}$ and $\text{train\_val}$ splits to conduct comprehensive evaluation and analysis on our method. We also conduct experiments on DAVIS17-RVOS \cite{khoreva2018video}. 
% For the evaluation metrics, we adopt the region similarity $\mathcal{J}$, contour similarity $\mathcal{F}$ and their average value
% $\mathcal{J} \& \mathcal{F}$ provided by \cite{seo2020urvos} to evaluate our method. We also evaluate our method with frame-level metrics: Precision $P@X$ and mean average precision (mAP) in this split for comprehensive analysis.

\subsection{Annotation Setting} \label{annotation setting}
In Table~\ref{sota}, we first compare the effectiveness of different annotation settings on SimRVOS. On average, a video contains 27.3 frames annotated with masks in YouTube-RVOS \cite{seo2020urvos}. Some methods \cite{cheng2022pointly, papadopoulos2017extreme} estimate that a polygon-based object mask and a bounding box annotation cost about 79 and 7 seconds, respectively. We can use them to estimate the time cost of different annotation settings. ``Full'' represents the fully-supervised setting, which costs about 2157 seconds to label the target object in a video. ``$\text{Weak}_{\text{B}}$'' denotes that all frames are annotated with bounding boxes, which is about $\times 11$ times faster than ``Full''.  We also attempt only annotating the frame where the target object first appears with a mask, which is denoted as ``$\text{Weak}_{\text{M}}$'' and is about $\times 27$ times faster than ``Full''. The proposed annotation scheme ``$\text{Weak}_{\text{M+B}}$'' is the combination of ``$\text{Weak}_{\text{B}}$'' and ``$\text{Weak}_{\text{M}}$'', resulting in $\times 8$ times ($\frac{2157}{1 \times 79 + 26.3 \times 7}=8.2$)  faster than labeling all frames with masks.% √% √

Obviously, ``Full'' achieves the best performance, which gains 44.7 on $\mathcal{J} \& \mathcal{F}$. SimRVOS with ``$\text{Weak}_{\text{B}}$'' only achieves 37.8 on $\mathcal{J} \& \mathcal{F}$, which has a large gap with the fully-supervised setting. That is mainly because the bounding boxes can only teach the model to locate the target object but can not provide fine-grained segmentation information. An example in Figure~\ref{figure1} (b) also proves this, where the model trained with only bounding box annotations tends to generate coarse segmentation masks. We find that ``$\text{Weak}_{\text{M}}$'' achieves better performance than ``$\text{Weak}_{\text{B}}$'', while the gap between it and the fully-supervised setting still can not be ignored. Although this scheme can help the model learn to segment the target object with precise masks, by learning from only one frame with annotation, the model may be difficult to handle the appearance changes of the target object and occlusion. The proposed annotation scheme ``$\text{Weak}_{\text{M+B}}$'' combines the advantages of these two schemes and achieves 42.7 on $\mathcal{J} \& \mathcal{F}$, which further bridge the performance gap with the fully-supervised setting. % √% √

We further demonstrate the difference between our RVOS with weak supervision and some related video object segmentation tasks in Table~\ref{vos}. Here, VOS and RVOS represent the semi-supervised video object segmentation \cite{caelles2017one, mei2021transvos, yang2021associating} and fully-supervised referring video object Segmentation \cite{vaswani2017attention, botach2022end, wu2022language, zhao2022modeling}, respectively. The semi-supervised Video object segmentation indicates that the first-frame mask is available during inference but still requires full supervision during training. In contrast, our scheme only needs the mask for the frame where the target object first appears and box annotations for other frames. Our annotation scheme achieves ×8 times faster than labeling all frames with masks in VOS and RVOS.

\subsection{Comparison with State-of-the-art Methods} \label{Comparison with State-of-the-art}
We compare our model with state-of-the-art methods on the official validation set of YouTube-RVOS \cite{seo2020urvos} in Table~\ref{sota}. Since the authors of ReferFormer \cite{wu2022language} do not provide the results without using image datasets, we adopt the code they provide and train it with official hyperparameters only on YouTube-RVOS \cite{seo2020urvos}. Results from ReferFormer \cite{wu2022language} with the joint training on image datasets can be found in Section~\ref{Experiments Based on ReferFormer}. ``YOFO'' \cite{li2022you} even relies heavily on joint training to achieve desirable performance. This indicates that the insufficient training data in YouTube-RVOS \cite{seo2020urvos} may hinder the performance of state-of-the-art methods, which encourages us to build larger RVOS datasets with weak annotations. When it comes to the proposed models, ``SimRVOS'' represents the proposed simple baseline for RVOS. For fair comparison, our model is also trained by 18 epochs. The ``SimRVOS+Ours'' indicates the model equipped with the proposed LGCFS and BLCL. Results illustrate that ``SimRVOS+Ours'' with weak supervision can even surpass ``SimRVOS'' with fuly supervision and achieve similar performance with other fully-supervised RVOS methods \eg URVOS \cite{seo2020urvos}, CMPC-V \cite{liu2021cross}, LBDT \cite{ding2022language}, MLRL \cite{wu2022multi}, and YOFO \cite{li2022you}. Without pretraining or joint training, the best method ReferFormer \cite{wu2022language} trained in full supervision and weak supervision achieves 49.9 and 48.9 on $\mathcal{J} \& \mathcal{F}$, respectively. When equipped with our methods, the performance of ``ReferFormer+Ours'' (Details can be found in Section~\ref{Experiments Based on ReferFormer}) can be further improved to 50.1 on $\mathcal{J} \& \mathcal{F}$ with $\text{Weak}_{\text{M+B}}$ supervision, which surpasses the original ReferFormer and other fully-supervised methods obviously. Both of these show the effectiveness of our methods.% √% √

% we adopt the code they provide and train it 18 epochs on YouTube-RVOS \cite{seo2020urvos}, which has the same total training epochs as their official settings.  
% we adopt the code they provide and train it with official settings on YouTube-RVOS \cite{seo2020urvos}, which has the same total training epochs as their official settings.

\section{Analysis}

\begin{table*}[t!]
	\centering

    \footnotesize
	\renewcommand{\arraystretch}{1.}
	\renewcommand{\tabcolsep}{1.5mm}
 
	\caption{Comparison of using different settings in LGCFS.}
  \vspace{-3mm}
	\begin{tabular}{cc|ccc|cccccc|cc} 

	\toprule 
	   \multicolumn{2}{c|}{\multirow{3}{*}{Name}}   &  \multicolumn{11}{c}{YouTube-RVOS train\_val}   \\   \cmidrule(lr){3-13} 
	   
	    \multicolumn{2}{c|}{}   &  \multicolumn{3}{c|}{Video Metrics} & \multicolumn{5}{c|}{Precision}  & \multicolumn{1}{c|}{mAP}    & \multicolumn{2}{c}{IoU} \\     
	                  
       \multicolumn{2}{c|}{}   & \multicolumn{1}{c}{$\mathcal{J} \& \mathcal{F}$} & \multicolumn{1}{c}{$\mathcal{J}$} & \multicolumn{1}{c|}{$\mathcal{F}$} & \multicolumn{1}{c}{P@0.5}               & \multicolumn{1}{c}{P@0.6} &\multicolumn{1}{c}{P@0.7}      & \multicolumn{1}{c}{P@0.8}  &\multicolumn{1}{c|}{P@0.9}  & \multicolumn{1}{c|}{0.5:0.95}    & \multicolumn{1}{c}{Over All} & \multicolumn{1}{c}{Mean} \\  \midrule

	\multicolumn{2}{c|}{w/o LGCFS}   & \multicolumn{1}{c}{61.8}  & \multicolumn{1}{c}{62.9}  & \multicolumn{1}{c|}{60.7}  & \multicolumn{1}{c}{49.8}   & \multicolumn{1}{c}{41.9}   & \multicolumn{1}{c}{33.4}   & \multicolumn{1}{c}{22.7} & \multicolumn{1}{c}{8.6}  & \multicolumn{1}{|c|}{28.6}      & \multicolumn{1}{c}{51.5}    & \multicolumn{1}{c}{46.6}  \\ \midrule

   \multicolumn{2}{c|}{first frame}   & \multicolumn{1}{c}{63.0}   & \multicolumn{1}{c}{64.0}   & \multicolumn{1}{c|}{62.0}  & \multicolumn{1}{c}{53.6}
   & \multicolumn{1}{c}{47.5}  & \multicolumn{1}{c}{38.6} & \multicolumn{1}{c}{26.9} & \multicolumn{1}{c|}{11.3} & \multicolumn{1}{c|}{32.7} & \multicolumn{1}{c}{54.3}    & \multicolumn{1}{c}{48.9}\\   \midrule
   
   \multicolumn{2}{c|}{first frame + others w/ avg}   & \multicolumn{1}{c}{63.4}   & \multicolumn{1}{c}{64.4}   & \multicolumn{1}{c|}{62.4} 
   & \multicolumn{1}{c}{53.8}  & \multicolumn{1}{c}{47.5}  & \multicolumn{1}{c}{38.7} & \multicolumn{1}{c}{27.4} & \multicolumn{1}{c|}{11.7} & \multicolumn{1}{c|}{32.9} & \multicolumn{1}{c}{54.8}    & \multicolumn{1}{c}{49.1}  \\  
   
   \multicolumn{2}{c|}{first frame + others w/o avg}   & \multicolumn{1}{c}{\textbf{63.9}}   & \multicolumn{1}{c}{\textbf{64.7}}   & \multicolumn{1}{c|}{\textbf{63.2}}   & \multicolumn{1}{c}{\textbf{56.0}} & \multicolumn{1}{c}{\textbf{49.2}} & \multicolumn{1}{c}{\textbf{40.8}} & \multicolumn{1}{c}{\textbf{28.5}} & \multicolumn{1}{c|}{\textbf{11.7}} & \multicolumn{1}{c|}{\textbf{34.3}}  & \multicolumn{1}{c}{\textbf{56.3}}    & \multicolumn{1}{c}{\textbf{50.7}}  \\

	\bottomrule
	\end{tabular}
        \label{LGCFS_ablation}
\end{table*}

% \begin{table}[t!]
% 	\centering
%     \footnotesize
% 	\renewcommand{\arraystretch}{1.0}
% 	\renewcommand{\tabcolsep}{3.6mm}
% 		\caption{Comparison of using different settings in LGCFS.}

% 	\begin{tabular}{cc|ccc|} 
% 	   	    \toprule 
% 	                  \multicolumn{2}{c|}{\multirow{2}{*}{Name}}   &  \multicolumn{3}{c}{YouTube-RVOS train\_val}  \\ \cmidrule(lr){3-5}

% 	               \multicolumn{2}{c|}{}  & \multicolumn{1}{|c}{$\mathcal{J} \& \mathcal{F}$} & \multicolumn{1}{c}{$\mathcal{J} $} & \multicolumn{1}{c}{$\mathcal{F}$}    \\   \midrule

%               \multicolumn{2}{c|}{w/o LGCFS}   & \multicolumn{1}{|c}{61.8}  & \multicolumn{1}{c}{62.9}  & \multicolumn{1}{c}{60.7}   \\   \midrule

% 	           \multicolumn{2}{c|}{first frame}   & \multicolumn{1}{|c}{63.0}   & \multicolumn{1}{c}{64.0}   & \multicolumn{1}{c}{62.0}  \\   \midrule
	           
% 	           \multicolumn{2}{c|}{first frame + others w/ avg}   & \multicolumn{1}{|c}{63.4}   & \multicolumn{1}{c}{64.4}   & \multicolumn{1}{c}{62.4} \\ 
	           
%    	           \multicolumn{2}{c|}{first frame + others w/o avg}   & \multicolumn{1}{|c}{\textbf{63.9}}   & \multicolumn{1}{c}{\textbf{64.7}}   & \multicolumn{1}{c}{\textbf{63.2}} \\  
	           
% 	       		       	\bottomrule
% 	\end{tabular}

% 	\label{LGCFS_ablation}
% 	% \vspace{-3mm}
% \end{table}

\begin{table*}[t!]
	\centering
    \footnotesize
	\renewcommand{\arraystretch}{1.0}
	\renewcommand{\tabcolsep}{1.5mm}
		\caption{Comparison of using different settings in BLCL.}
  \vspace{-3mm}
% 	"Language-Vision": using language-vision contrast. "Consistency": using consistency contrast. "Pseudo": using pseudo labels for frames only with bounding box annotations.
	\begin{tabular}{cccccc|ccc|cccccc|cc} 
	   \toprule 
	     \multicolumn{6}{c|}{\multirow{3}{*}{Setting}}  &  \multicolumn{11}{c}{YouTube-RVOS train\_val} \\ \cmidrule(lr){7-17}

	    \multicolumn{6}{c|}{}   &  \multicolumn{3}{c|}{Video Metrics} & \multicolumn{5}{c|}{Precision}  & \multicolumn{1}{c|}{mAP}    & \multicolumn{2}{c}{IoU} \\

	   \multicolumn{2}{c}{Language-Vision}  &  \multicolumn{2}{c}{Consistency}  &  \multicolumn{2}{c|}{Pseudo}   & \multicolumn{1}{c}{$\mathcal{J} \& \mathcal{F}$} & \multicolumn{1}{c}{$\mathcal{J} $} & \multicolumn{1}{c|}{$\mathcal{F}$}  & \multicolumn{1}{c}{P@0.5}               & \multicolumn{1}{c}{P@0.6} &\multicolumn{1}{c}{P@0.7}      & \multicolumn{1}{c}{P@0.8}  &\multicolumn{1}{c|}{P@0.9}  & \multicolumn{1}{c|}{0.5:0.95}    & \multicolumn{1}{c}{Over All} & \multicolumn{1}{c}{Mean}\\  	 \midrule
	               
        \multicolumn{2}{c}{} & \multicolumn{2}{c}{} & \multicolumn{2}{c|}{}  & \multicolumn{1}{c}{63.9}   & \multicolumn{1}{c}{64.7}   & \multicolumn{1}{c|}{63.2} & \multicolumn{1}{c}{56.0} & \multicolumn{1}{c}{49.2} & \multicolumn{1}{c}{40.8} & \multicolumn{1}{c}{28.5} & \multicolumn{1}{c}{11.7} & \multicolumn{1}{|c|}{34.3}   & \multicolumn{1}{c}{56.3}    & \multicolumn{1}{c}{50.7}  \\   \midrule
	               
       \multicolumn{2}{c}{\ding{51}} & \multicolumn{2}{c}{} & \multicolumn{2}{c|}{}  & \multicolumn{1}{c}{64.4}   & \multicolumn{1}{c}{65.2}   & \multicolumn{1}{c|}{63.7} & \multicolumn{1}{c}{56.2} & \multicolumn{1}{c}{49.9} & \multicolumn{1}{c}{41.9} & \multicolumn{1}{c}{29.4} & \multicolumn{1}{c|}{12.3} & \multicolumn{1}{c|}{35.1} & \multicolumn{1}{c}{55.3} & \multicolumn{1}{c}{50.8} \\

      \multicolumn{2}{c}{\ding{51}} & \multicolumn{2}{c}{} & \multicolumn{2}{c|}{\ding{51}}  & \multicolumn{1}{c}{64.6}   & \multicolumn{1}{c}{65.5} & \multicolumn{1}{c|}{63.7}  & \multicolumn{1}{c}{56.3} & \multicolumn{1}{c}{50.6} & \multicolumn{1}{c}{42.4} & \multicolumn{1}{c}{29.9} \ & \multicolumn{1}{c|}{12.6} & \multicolumn{1}{c|}{35.4} & \multicolumn{1}{c}{55.7} & \multicolumn{1}{c}{51.0} \\   \midrule

      \multicolumn{2}{c}{} & \multicolumn{2}{c}{\ding{51}} & \multicolumn{2}{c|}{}  & \multicolumn{1}{c}{64.5}  & \multicolumn{1}{c}{65.3}   & \multicolumn{1}{c|}{63.6}  & \multicolumn{1}{c}{56.1} & \multicolumn{1}{c}{49.2} & \multicolumn{1}{c}{41.2} & \multicolumn{1}{c}{29.8} \ & \multicolumn{1}{c|}{12.6} & \multicolumn{1}{c|}{34.8} & \multicolumn{1}{c}{56.0} & \multicolumn{1}{c}{50.6} \\  

    \multicolumn{2}{c}{} & \multicolumn{2}{c}{\ding{51}} & \multicolumn{2}{c|}{\ding{51}}  & \multicolumn{1}{c}{64.7}   & \multicolumn{1}{c}{65.5} & \multicolumn{1}{c|}{63.9}  & \multicolumn{1}{c}{56.3} & \multicolumn{1}{c}{49.7} & \multicolumn{1}{c}{41.7}  & \multicolumn{1}{c}{30.3} & \multicolumn{1}{c|}{12.8} & \multicolumn{1}{c|}{35.2} & \multicolumn{1}{c}{56.3} & \multicolumn{1}{c}{50.8} \\  \midrule

    \multicolumn{2}{c}{\ding{51}} & \multicolumn{2}{c}{\ding{51}} & \multicolumn{2}{c|}{}  & \multicolumn{1}{c}{64.7}   & \multicolumn{1}{c}{65.6} & \multicolumn{1}{c|}{63.9}  & \multicolumn{1}{c}{56.9}  & \multicolumn{1}{c}{50.3}  & \multicolumn{1}{c}{41.8} & \multicolumn{1}{c}{29.6} & \multicolumn{1}{c|}{12.8} & \multicolumn{1}{c|}{35.2} & \multicolumn{1}{c}{56.5} & \multicolumn{1}{c}{51.2} \\

    \multicolumn{2}{c}{\ding{51}} & \multicolumn{2}{c}{\ding{51}} & \multicolumn{2}{c|}{\ding{51}}    &  \multicolumn{1}{c}{\textbf{65.1}}   & \multicolumn{1}{c}{\textbf{66.0}} & \multicolumn{1}{c|}{\textbf{64.2}} & \multicolumn{1}{c}{\textbf{57.4}} & \multicolumn{1}{c}{\textbf{51.2}} & \multicolumn{1}{c}{\textbf{42.9}} & \multicolumn{1}{c}{\textbf{30.5}} & \multicolumn{1}{c|}{\textbf{12.7}} & \multicolumn{1}{c|}{\textbf{35.8}}  & \multicolumn{1}{c}{\textbf{56.7}}    & \multicolumn{1}{c}{\textbf{51.6}} \\  	       	\bottomrule

	\end{tabular}

	\label{BLCL_ablation}
	% \vspace{-7mm}
\end{table*}

\subsection{Ablation Study of Model Design} \label{ablation study}
In this section, we conduct experiments to verify the effectiveness of each component in our model.

\noindent\textbf{Effectiveness of Each Component.}
We verify the effectiveness of each component on the official validation sets in Table~\ref{ablation_table_official}. The setting is consistent with Section~\ref{Comparison with State-of-the-art}. On YouTube-RVOS \cite{seo2020urvos}, the proposed LGCFS and BLCL can improve the performance gradually, from 41.7 to 45.6  on $\mathcal{J}$. On DAVIS17-RVOS \cite{khoreva2018video}, ``SimRVOS+LGCFS'' improves the performance dramatically, from 40.8 to 43.5 on $\mathcal{J}$ and ``SimRVOS+LGCFS+BLCL'' achieves better performance. These prove the generalization ability of the proposed methods.

In Table~\ref{ablation_table_splitted}, we also conduct experiments on the $\text{train\_val}$ split to comprehensively analyze each component on both video metrics and frame metrics. For these experiments, models are trained on $\text{train\_train}$ split by 6 epochs in total. This schedule is as same as ReferFormer \cite{wu2022language}. With the proposed LGCFS, ``SimRVOS+LGCFS'' improve the performance by a large margin on all metrics \eg from 61.8 to 63.9 on $\mathcal{J} \& \mathcal{F}$. The proposed method also enhances the frame-level metrics, such as mAP, which increases from 28.6 to 34.3. This benefits from that LGCFS not only leverages the frame with mask annotations to supervise the learning of other frames but also fully takes advantage of abundant bounding box annotations. Comparing ``SimRVOS+LGCFS'' and ``SimRVOS+LGCFS+BLCL'', we find that the performance can be further improved with the enhanced ability of foreground-background discrimination from the proposed BLCL. Both the video metric $\mathcal{J} \& \mathcal{F}$ and the frame-level metric mAP are enhanced by the proposed BLCL, achieving 65.1 and 35.8, respectively, compared to 63.9 and 34.3 of ``SimRVOS+LGCFS''. These results verify the superiority of the proposed LGCFS and BLCL.

% Results``LGCFS" and`BLCL" can also improve the frame-level metrics \eg mAP, Precision, and IoU consistently.

\begin{figure*}[!t]
  \centering
  \includegraphics[width=1.0\linewidth]{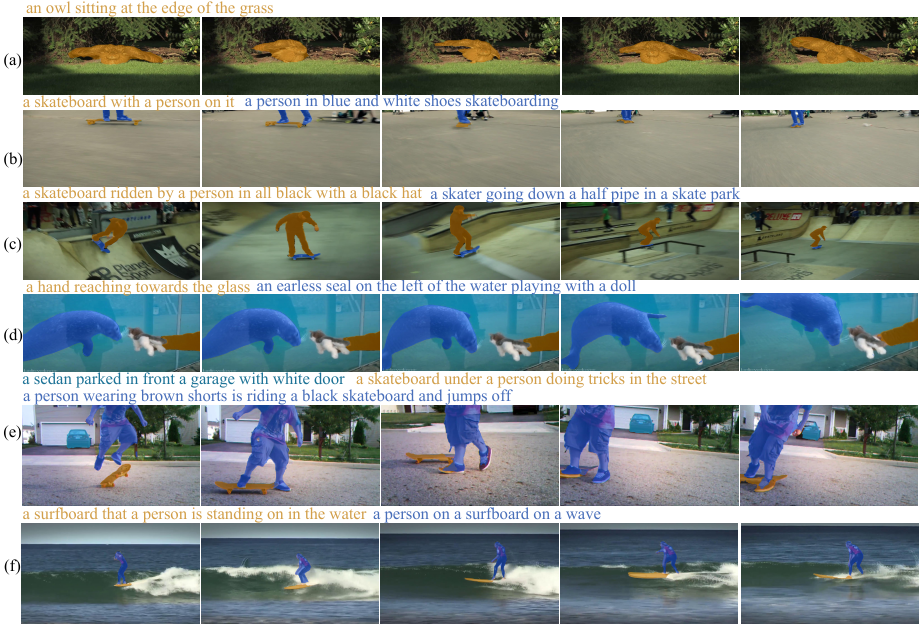}
  % \vspace{-6mm}
    \vspace{-5mm}
  \caption{Qualitative results from ``SimRVOS+Ours'' in Table~\ref{ablation_table_official}. All examples here are sampled from the official validation set of YouTube-RVOS \cite{seo2020urvos}.}
  \label{qualitative}
  	% \vspace{-5mm}
\end{figure*}

\begin{figure}[!t]
    \centering
    \includegraphics[width=1\linewidth]{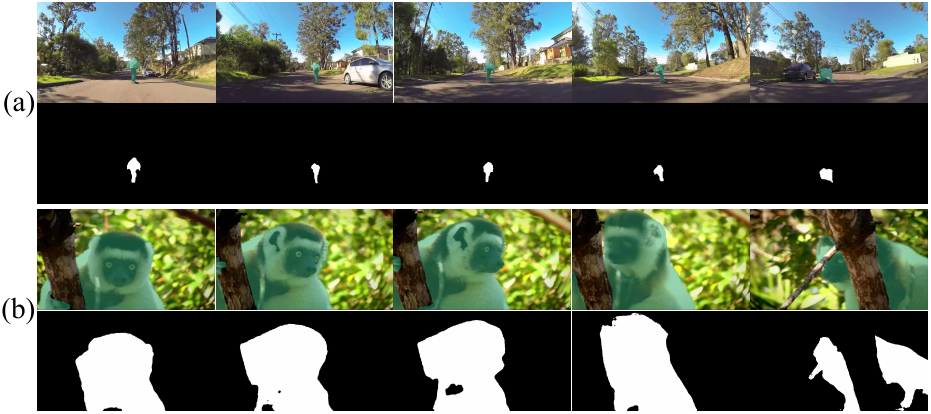}
    \vspace{-5mm}
    \caption{We adopt the language-guided dynamic filters from the frame in the first column to predict the masks for other frames. First row: the ground-truth masks. Second row: our predictions.} 
    \label{LGCFS_visualize}
% \vspace{-4mm}
\end{figure} % √

\begin{figure}[!t]
    \centering
    \includegraphics[width=1\linewidth]{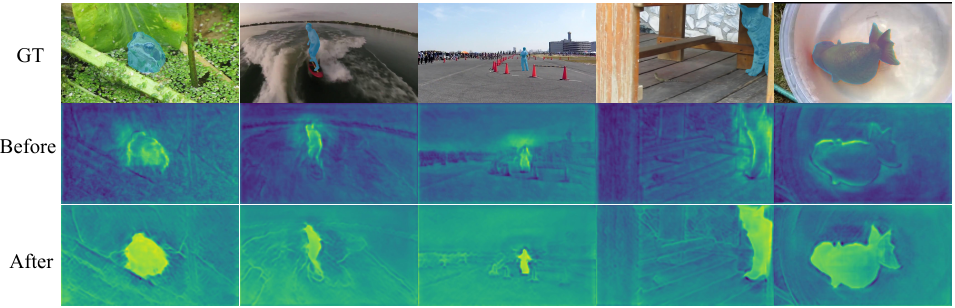}
    \vspace{-5mm}
    \caption{We visualize the feature maps before and after the BLCL by meaning them along the channel dimension.}  % √
    \label{BLCL_visualize}
    % \vspace{-2mm}
\end{figure}

\noindent\textbf{Different Settings in LGCFS.}
To thoroughly analyze the effectiveness of LGCFS, we further conduct ablation studies on it, whose results are shown in Table~\ref{LGCFS_ablation}. ``w/o LGFCS'' means that no LGCFS loss is adopted, which is the same as ``SimRVOS'' in Table~\ref{ablation_table_splitted}. ``first frame'' means that we only adopt the language-guided dynamic filters from other frames to generate masks for the frame with mask annotation. Hence, only Equation~\ref{seg loss} is adopted.
Compared with ``w/o LGFCS'', the performance of ``first frame'' is significantly improved in all metrics, \eg $\mathcal{J} \& \mathcal{F}$ from 61.8 to 63.0. This demonstrates the superiority of LGCFS. ``first frame + others w/ avg'' means that both Equation~\ref{seg loss} and ~\ref{MIL loss} are adopted, but the MIL losses from frames with bonding box annotations are averaged by the number of these frames, which means that these MIL losses are given a smaller weight. ``first frame + others w/o avg'' means that no average is adopted. The performance is slightly improved by ``first frame + others w/ avg'', which achieves 63.4 $\mathcal{J} \& \mathcal{F}$ and 32.9 mAP, respectively, over 63.0 and 32.7 of ``first frame''. A further improvement is obtained by ``first frame + others w/o avg'', which reaches 63.9 $\mathcal{J} \& \mathcal{F}$ and 34.3 mAP, respectively. This demonstrates the importance of the supervision from bounding boxes. Therefore, we adopt first frame + others w/o avg'' as our setting for LGCFS.

In Figure~\ref{LGCFS_visualize}, we visualize two examples from the train\_val split, where the language-guided dynamic filters from the frame in the first column are adopted to predict the masks for all frames. The appearance of the target object in (a) changes largely while the target object in (b) is occluded. Results show that the learned language-guided dynamic filter in one frame is robust enough to handle these challenging scenes and segment the object in other frames accurately. This confirms that LGCFS enables the model to produce fine-grained masks with weak supervision and to handle the appearance variations of the target object.

% √ % While, ``first frame + others w/o avg" can further improve $\mathcal{J} \& \mathcal{F}$ and mAP to 63.9 and 34.3, respectively. This illustrates that the supervision from bounding boxes is essential. Hence, we choose ``first frame + others w/o avg"  as our setting for LGCFS. 

\noindent\textbf{Different Settings in BLCL.}
We also conduct experiments to further explore different settings of BLCL in Table~\ref{BLCL_ablation}. ``Language-Vision'' means that we only adopt the language-vision contrastive loss $\mathcal{L}_{\text{LV}}$. ``Consistency'' represents only the consistency contrast $\mathcal{L}_{\text{CC}_{fg}} + \mathcal{L}_{\text{CC}_{bg}}$ is adopted. ``Pseudo'' indicates that we use the predicted masks as pseudo labels for frames without mask annotations to distinguish foreground and background features with the aid of bounding annotations. Without ``Pseudo'', we use bounding boxes to find background features outside the bounding box accurately for those frames only with bounding boxes. This implies that, without ``Pseudo'', fewer positive and negative samples are available during training. Comparing the first row with the second and $4$-th rows, we can find that both language-vision contrast and consistency contrast can improve performance obviously in most metrics. The $6$-th row shows that combining these two loss functions can further improve performance. Results also demonstrate that models with ``Pseudo'' always beat those models without ``Pseudo''. That is mainly because ``Pseudo'' can provide more positive and negative samples in contrastive learning, which makes the model learn more robust foreground-background discrimination. Thus, we adopt the combination of two losses with ``Pseudo'' as the final setting for BLCL. % √

To show the source of performance gain from BLCL,  we visualize the feature map before BLCL $\mathbf{F}_{\text{FPN}}^t$ and the feature after BLCL $\mathbf{F}_{\text{ENH}}^t$ in Figure~\ref{BLCL_visualize}. We can find that BLCL makes foreground features more distinguishable from background features. This show that the improved pixel-level discriminative representation contributes to the performance gain from BLCL.

% ``Pseudo" means that we adopt the predicted masks as pseudo labels for frames without mask annotations to separate foreground and background features with the help of bounding annotations. Note that when we do not adopt ``Pseudo", bounding boxes are adopted to separate background features out of the bounding box precisely for those frames only with bounding boxes. This means that, without ``Pseudo", less foreground and background samples are available during training. 

\subsection{Qualitative Results}
Figure~\ref{qualitative} shows some predictions from our model ``SimRVOS+LGCFS+BLCL'' for some challenging samples. For instance, the owl in (a) has a similar color to the background, the person in (b) is incomplete, the appearance and viewpoint vary significantly in (c), and the sedan is absent in some frames in (e). By learning from both mask and bounding box annotations effectively, our model can locate and segment the target object accurately in these challenging scenarios.

\subsection{Experiments Based on ReferFormer} \label{Experiments Based on ReferFormer}
Our method is also compatible with the state-of-the-art method ReferFormer \cite{wu2022language}, which can be viewed as a stronger baseline. We adopt the query whose prediction is matched with the ground truth to conduct our LGCFS and BLCL, denoted as ``ReferFormer+Ours''. The experimental results are shown in Table~\ref{ReferFormer}. In the fully-supervised setting, the performance of ReferFormer with our LGCFS and BLCL is obviously improved compared with the original version. We are surprised to find that the ReferFormer with our LGCFS and BLCL trained with our weak annotation can even surpass the original full-supervised ReferFormer and achieves 50.1 $\mathcal{J} \& \mathcal{F}$, which further proves the superiority of our methods.

We also conduct experiments in the joint training setting, where we mix YouTube-RVOS \cite{seo2020urvos} and pseudo videos generated from RefCOCO \cite{yu2016modeling}, RefCOCOg \cite{yu2016modeling}, and RefCOCO+ \cite{mao2016generation} to training our model. This strategy is consistent with ReferFormer \cite{wu2022language}, and details can be found in their paper \cite{wu2022language}. ``Full'' means that both pseudo videos and YouTube-RVOS are in the fully-supervised setting. We find that ``ReferFormer+Ours'' can improve the $\mathcal{J} \& \mathcal{F}$ from 58.7 to 59.3. We also try to mix the YouTube-RVOS with our weak annotation scheme and fully-annotated pseudo videos to train the model, which is denoted as ``Mix''. It achieves 58.0 on $\mathcal{J} \& \mathcal{F}$, which is slightly lower than the fully-supervised setting. This promises that future works can leverage existing fully-supervised referring video or image object segmentation datasets and incoming weakly-annotated RVOS datasets to achieve super performance. 

We further evaluate “ReferFormer+Ours” against two models, YOFO \cite{li2022you} and OnlineRefer \cite{wu2023onlinerefer}, that are trained on both pseudo videos and YouTube-RVOS \cite{seo2020urvos}. Our model outperforms them even with “Mix” supervision. With full supervision, the performance gap between “ReferFormer+Ours” and OnlineRefer widens, \eg 59.3 vs 57.3 on $\mathcal{J} \& \mathcal{F}$. This demonstrates the effectiveness of our weak annotation scheme and proposed methods.

\begin{table}[t!]
	\centering
    % \footnotesize
	\caption{Experiments based on ReferFormer \cite{wu2022language}. ``Joint'': We use both pseudo videos from image datasets and real videos from YouTube-RVOS \cite{seo2020urvos}. ``Full'': Full supervision for training data. ``Weak'': Our weak annotation scheme for training data. ``Mix'': YouTube-RVOS with our weak annotation scheme and fully-annotated pseudo videos to train the model. $\dagger$ indicates that the model is first trained on pseudo videos and then finetuned on YouTube-RVOS \cite{seo2020urvos}.}
   \vspace{-3mm}
\setlength{\tabcolsep}{1.5mm}{
	\begin{tabular}{cc|c|c|c|ccc|} 

	   	    \toprule 
	                  \multicolumn{2}{c|}{\multirow{2}{*}{Name}} & \multicolumn{1}{c|}{\multirow{2}{*}{Supervision}}  & \multicolumn{1}{c|}{\multirow{2}{*}{Extra data}} &  \multicolumn{3}{c}{YouTube-RVOS val}  \\ 

	               \multicolumn{2}{c|}{} & \multicolumn{1}{c|}{}  & \multicolumn{1}{c|}{}  & \multicolumn{1}{c}{$\mathcal{J} \& \mathcal{F}$} & \multicolumn{1}{c}{$\mathcal{J}$}  & \multicolumn{1}{c}{$\mathcal{F}$} \\  \hline

   	       	         \multicolumn{2}{c|}{YOFO \cite{li2022you}}    & \multicolumn{1}{c|}{Full}    & \multicolumn{1}{c|}{Joint}  & \multicolumn{1}{c}{48.6} & \multicolumn{1}{c}{47.5}      & \multicolumn{1}{c}{49.7}    \\           
     
   	    \multicolumn{2}{c|}{OnlineRefer $\dagger$ \cite{wu2023onlinerefer}}    & \multicolumn{1}{c|}{Full}   & \multicolumn{1}{c|}{Joint}  & \multicolumn{1}{c}{57.3} & \multicolumn{1}{c}{55.6}      & \multicolumn{1}{c}{58.9}   \\ \midrule

   	       	 \multicolumn{2}{c|}{ReferFormer}    & \multicolumn{1}{c|}{Full}   & \multicolumn{1}{c|}{/}  & \multicolumn{1}{c}{49.9} & \multicolumn{1}{c}{49.2}      & \multicolumn{1}{c}{50.6}    \\

   	       	\multicolumn{2}{l|}{ReferFormer+Ours}      & \multicolumn{1}{c|}{Weak}  & \multicolumn{1}{c|}{/}   & \multicolumn{1}{c}{50.1}    & \multicolumn{1}{c}{48.9}      & \multicolumn{1}{c}{51.2}    \\

  	       	   \multicolumn{2}{c|}{ReferFormer+Ours}    & \multicolumn{1}{c|}{Full}   & \multicolumn{1}{c|}{/}  & \multicolumn{1}{c}{\textbf{50.6}} & \multicolumn{1}{c}{\textbf{49.8}}      & \multicolumn{1}{c}{\textbf{51.4}}   \\ \midrule

   	       	  \multicolumn{2}{c|}{ReferFormer}    & \multicolumn{1}{c|}{Full}   & \multicolumn{1}{c|}{Joint}  & \multicolumn{1}{c}{58.7} & \multicolumn{1}{c}{57.4}      & \multicolumn{1}{c}{60.1}   \\ 

  	   	      \multicolumn{2}{c|}{ReferFormer+Ours}    & \multicolumn{1}{c|}{Mix}   & \multicolumn{1}{c|}{Joint}  & \multicolumn{1}{c}{58.0} & \multicolumn{1}{c}{56.6}      & \multicolumn{1}{c}{59.3}   \\

   	       	\multicolumn{2}{c|}{ReferFormer+Ours}    & \multicolumn{1}{c|}{Full}   & \multicolumn{1}{c|}{Joint}  & \multicolumn{1}{c}{\textbf{59.3}} & \multicolumn{1}{c}{\textbf{58.0}}      & \multicolumn{1}{c}{\textbf{60.6}}   \\

	       	\bottomrule
	\end{tabular}}
	\label{ReferFormer}
 \vspace{-3mm}
\end{table}

\subsection{Influence of the Temporal Position of the Mask} \label{Influence of the Temporal Position of the Mask}
To investigate the influence of the temporal position of the frame with the mask, we learn ``SimRVOS+Ours” by randomly labeling one frame with a mask instead of always labeling the frame where the target object first appears with a mask. Other frames still are labeled with bounding boxes. From Table~\ref{temporal position}, we repeat the experiments three times and we find that the averaged result is 46.5 on $\mathcal{J} \& \mathcal{F}$, which is similar to ours (46.6). This means that the temporal position has little influence on performance, while we believe that labeling the frame where the target object first appears is easier and straightforward.

\begin{table}[t!]
	\centering
    \footnotesize
    \caption{Influence of the temporal position of the mask.}
    \vspace{-3mm}
\setlength{\tabcolsep}{4.5mm}{
	\begin{tabular}{lc|c|ccc|} 

\toprule 
\multicolumn{2}{l|}{\multirow{2}{*}{Name}}   &  \multicolumn{3}{c}{YouTube-RVOS val}  \\ 

\multicolumn{2}{l|}{} & \multicolumn{1}{c}{$\mathcal{J} \& \mathcal{F}$} & \multicolumn{1}{c}{$\mathcal{J}$}  & \multicolumn{1}{c}{$\mathcal{F}$} \\  
\midrule

\multicolumn{2}{l|}{SimRVOS+Ours}       & \multicolumn{1}{c}{46.6}    & \multicolumn{1}{c}{45.6}      & \multicolumn{1}{c}{47.6}    \\  \midrule

\multicolumn{2}{l|}{SimRVOS+Ours 1st-try}     & \multicolumn{1}{c}{46.7}    & \multicolumn{1}{c}{45.7}      & \multicolumn{1}{c}{47.6}    \\  

\multicolumn{2}{l|}{SimRVOS+Ours 2nd-try}      & \multicolumn{1}{c}{46.1}    & \multicolumn{1}{c}{45.1}      & \multicolumn{1}{c}{47.1}    \\  

\multicolumn{2}{l|}{SimRVOS+Ours 3rd-try}    & \multicolumn{1}{c}{46.7}    & \multicolumn{1}{c}{45.8}      & \multicolumn{1}{c}{47.6}    \\  \midrule

\multicolumn{2}{l|}{SimRVOS+Ours average}      & \multicolumn{1}{c}{46.5}    & \multicolumn{1}{c}{45.5}      & \multicolumn{1}{c}{47.4}    \\  \bottomrule
\end{tabular}}

\label{temporal position}
\vspace{-3mm}
\end{table}

\subsection{Importance of Cross-modal Fusion Module} \label{Ablation study on Cross-modal Fusion Module}
$\mathcal{A}_{\text{L2V}}$ and $\mathcal{A}_{\text{V2L}}$ are main components of cross-modal fusion modules. Removing $\mathcal{A}_{\text{V2L}}$ is not feasible since it encodes the visual representation of each frame to obtain frame-wise dynamic filters $f_{\theta}^t$, which leads to terrible performance. We conduct an experiment by removing $\mathcal{A}_{\text{L2V}}$ from ``SimRVOS'' and find the performance drops from 42.7 to 41.8 on $\mathcal{J} \& \mathcal{F}$, which indicates that linguistic feature is helpful to improve the visual representation. Results are shown in Table~\ref{remove l2v}.

\begin{table}[t!]
\centering
\footnotesize
\caption{Importance of on Cross-modal Fusion Module.}
\vspace{-3mm}
\setlength{\tabcolsep}{5.0mm}{
\begin{tabular}{lc|ccc|} 
\toprule 
\multicolumn{2}{l|}{\multirow{2}{*}{\textbf{Method}}}  &  \multicolumn{3}{c}{YouTube-RVOS val}  \\ 

   \multicolumn{2}{l|}{}   & \multicolumn{1}{c}{$\mathcal{J} \& \mathcal{F}$} & \multicolumn{1}{c}{$\mathcal{J}$}  & \multicolumn{1}{c}{$\mathcal{F}$} \\  
   \midrule

\multicolumn{2}{l|}{SimRVOS}      & \multicolumn{1}{c}{42.7}    & \multicolumn{1}{c}{41.7}      & \multicolumn{1}{c}{43.6}    \\   

\multicolumn{2}{l|}{SimRVOS w/o $\mathcal{A}_{\text{L2V}}$}      & \multicolumn{1}{c}{41.8}    & \multicolumn{1}{c}{41.0}      & \multicolumn{1}{c}{42.7}    \\   

\bottomrule      
\end{tabular}}
\label{remove l2v}
\end{table}

\subsection{Pre-defined Threshold $d_{\mathrm{th}}$ in BLCL} \label{Comparison with a Two-Step Method}
We conduct experiments to investigate the influence of the threshold $d_{\mathrm{th}}$ in BLCL. Table~\ref{threshold} shows the results of sweeping $d_{\mathrm{th}}$ from 0.6 to 0.9. We found that our model is not sensitive to the threshold. As $d_{\mathrm{th}}$ increases, BLCL can be trained with more accurate but fewer samples. We adopted $d_{\mathrm{th}}=0.9$ as the threshold in BLCL for our model since it provides the most accurate training samples.

\begin{table}[t!]
	\centering
    \footnotesize
    \caption{Influence of the pre-defined threshold $d_{\mathrm{th}}$ in BLCL $d_{\mathrm{th}}$ .}
    \vspace{-3mm}
\setlength{\tabcolsep}{7.0mm}{
	\begin{tabular}{lc|c|ccc|} 

\toprule 
\multicolumn{2}{l|}{\multirow{2}{*}{Threshold}}   & \multicolumn{3}{c}{YouTube-RVOS val}  \\ 

\multicolumn{2}{l|}{} & \multicolumn{1}{c}{$\mathcal{J} \& \mathcal{F}$} & \multicolumn{1}{c}{$\mathcal{J}$}  & \multicolumn{1}{c}{$\mathcal{F}$} \\  
\midrule

% \multicolumn{2}{l|}{0.5}     & \multicolumn{1}{c}{45.2}    & \multicolumn{1}{c}{44.1}      & \multicolumn{1}{c}{46.3}    \\   
\multicolumn{2}{l|}{0.6}     & \multicolumn{1}{c}{46.8}    & \multicolumn{1}{c}{45.6}      & \multicolumn{1}{c}{47.9}    \\   
\multicolumn{2}{l|}{0.7}     & \multicolumn{1}{c}{47.2}    & \multicolumn{1}{c}{46.2}      & \multicolumn{1}{c}{48.2}    \\   
\multicolumn{2}{l|}{0.8}     & \multicolumn{1}{c}{47.3}    & \multicolumn{1}{c}{46.3}      & \multicolumn{1}{c}{48.4}    \\   
\multicolumn{2}{l|}{0.9}     & \multicolumn{1}{c}{46.6}    & \multicolumn{1}{c}{45.6}      & \multicolumn{1}{c}{47.6}    \\   \bottomrule

\end{tabular}}

\label{threshold}
\vspace{-3mm}
\end{table}

\subsection{Experiments with Different Backbones} \label{Experiments with different backbones}
In Table~\ref{SimRVOS+Ours with different backbones}, we replace the ResNet-50 backbone in ``SimRVOS+Ours'' with a stronger 2D backbone ResNet-101 \cite{he2016deep} and a 3D backbone VideoSwin-Tiny \cite{liu2022video}. The results show that our method’s performance consistently improves with a stronger backbone.

\begin{table}[t!]
	\centering
    \footnotesize
    \caption{SimRVOS+Ours with different backbones.}
    \vspace{-3mm}
\setlength{\tabcolsep}{2.5mm}{
	\begin{tabular}{lc|c|c|ccc|} 
	    \toprule 
	          \multicolumn{2}{l|}{\multirow{2}{*}{\textbf{Method}}} & \multicolumn{1}{c|}{\multirow{2}{*}{\textbf{Backbone}}}  &  \multicolumn{3}{c}{YouTube-RVOS val}  \\ 

	               \multicolumn{2}{l|}{} & \multicolumn{1}{c|}{}   & \multicolumn{1}{c}{$\mathcal{J} \& \mathcal{F}$} & \multicolumn{1}{c}{$\mathcal{J}$}  & \multicolumn{1}{c}{$\mathcal{F}$} \\  
	               \midrule

            \multicolumn{2}{l|}{SimRVOS+Ours}      & \multicolumn{1}{c|}{ResNet-50 \cite{he2016deep}}  & \multicolumn{1}{c}{46.6}    & \multicolumn{1}{c}{45.6}      & \multicolumn{1}{c}{47.6}    \\   
     	
      	\multicolumn{2}{l|}{SimRVOS+Ours}      & \multicolumn{1}{c|}{ResNet-101 \cite{he2016deep}}  &  \multicolumn{1}{c}{47.3}    & \multicolumn{1}{c}{45.9}      & \multicolumn{1}{c}{48.7}   \\    
      	
      	\multicolumn{2}{l|}{SimRVOS+Ours}      & \multicolumn{1}{c|}{VideoSwin-Tiny \cite{liu2022video}}   &  \multicolumn{1}{c}{50.2}    & \multicolumn{1}{c}{48.9}      & \multicolumn{1}{c}{51.4}    \\  \bottomrule

	\end{tabular}}
	\label{SimRVOS+Ours with different backbones}
 \vspace{-3mm}
\end{table}

\subsection{Comparison with a Two-Step Method} \label{Comparison with a Two-Step Method}
We compare our method with a straightforward two-step method, which first adopts the first-frame mask and bounding box annotations to train a model, then uses pseudo masks generated from the trained model to train the model in a fully-supervised manner. We adopt $\mathcal{L}_{\text{MASK}}$ and $\mathcal{L}_{\text{MIL}}$ as the loss functions for the frame with the mask and bounding box annotation, respectively. We conduct experiments on ReferFormer \cite{wu2022language} and reulst are shown in Table~\ref{two-step}. We observe that the performance drops from 50.1 (``ReferFormer+Our'') to 49.0 (``ReferFormer+Two-Step'') on $\mathcal{J} \& \mathcal{F}$. The performance of ``ReferFormer+Two-Step'' is slightly better than ``ReferFormer'', while our method surpasses ``ReferFormer'' by a large margin, which means that the two-step method is less effective in RVOS with weak supervision. We believe that our end-to-end learning paradigm improves the model's robustness against the label noise, while the two-step method suffers from error accumulation in the first step.

\begin{table}[t!]
\centering
\footnotesize
\caption{Comparison with a two-step method.}
    \vspace{-3mm}
\setlength{\tabcolsep}{4.5mm}{
\begin{tabular}{cc|c|ccc|} 
%\hline
\toprule 
\multicolumn{2}{c|}{\multirow{2}{*}{Name}}   &  \multicolumn{3}{c}{YouTube-RVOS val}  \\ 

\multicolumn{2}{c|}{} & \multicolumn{1}{c}{$\mathcal{J} \& \mathcal{F}$} & \multicolumn{1}{c}{$\mathcal{J}$}  & \multicolumn{1}{c}{$\mathcal{F}$} \\  
\midrule

\multicolumn{2}{l|}{ReferFormer \cite{wu2022language}}        & \multicolumn{1}{c}{48.9} & \multicolumn{1}{c}{47.9}      & \multicolumn{1}{c}{49.9}    \\ \midrule

\multicolumn{2}{l|}{ReferFormer+Two-Step}  & \multicolumn{1}{c}{49.0}    & \multicolumn{1}{c}{48.1}      & \multicolumn{1}{c}{49.9}    \\   

\multicolumn{2}{l|}{ReferFormer+Ours}      & \multicolumn{1}{c}{\textbf{50.1}}    & \multicolumn{1}{c}{\textbf{48.9}}      & \multicolumn{1}{c}{\textbf{51.2}}    \\   
\bottomrule
\end{tabular}}

\label{two-step}
\end{table}

\subsection{Results on A2D and JHMDB} \label{Results on A2D and JHMDB}
We also conduct experiments on A2D-Sentences \cite{gavrilyuk2018actor} and JHMDB-Sentences \cite{gavrilyuk2018actor} in Table~\ref{A2D} and Table~\ref{JHMDB}, respectively. All models adopt the training set of A2D-Sentences to train and are evaluated on the validation set of A2D-Sentences and the whole JHMDB-Sentences. In A2D-Sentences, 3-5 frames are annotated with masks in each video. Most methods here adopt 3D encoders \eg VideoSwin-Tiny \cite{liu2022video} as the backbone to leverage a large number of frames without annotations, which promotes the performance largely. To show the effectiveness of our LGCFS and BLCL, we ignore frames without annotation and only adopt the mask from the frame where the target object first appears and bounding boxes from rest frames to train ``SimRVOS+Ours'' with ResNet-50 \cite{he2016deep}. Although this may hinder our performance, our model still outperforms fully-supervised methods before 2022 on most metrics. Note that only our method and \cite{chen2022weakly} are trained without full supervision. Since Chen \etal \cite{chen2022weakly} do not use any fine-grained annotations during training, our method surpasses theirs by a large margin easily.

\begin{table*}[t!]
	\centering
	\scriptsize
	\renewcommand{\arraystretch}{1.0}
	\renewcommand{\tabcolsep}{4.3mm}
	\caption{Comparison with state-of-the-art methods on A2D-Sentences testing set. $\dagger$ denotes adopting additional optical flow input. * denotes that the model is trained with weak annotations or without annotations.}
	\begin{tabular}{cccc|cc|ccccc|c|cc} 
	    \toprule 
        \multicolumn{4}{c|}{\multirow{2}{*}{Methods}}  & \multicolumn{2}{c|}{\multirow{2}{*}{Year}} & \multicolumn{5}{c|}{Precision}  & \multicolumn{1}{c|}{mAP}  & \multicolumn{2}{c}{IoU} \\ %\cline{7-14}
        
        \multicolumn{4}{c|}{}                                   &  \multicolumn{2}{c|}{}                                 &\multicolumn{1}{c}{P@0.5}               &\multicolumn{1}{c}{P@0.6} &\multicolumn{1}{c}{P@0.7}      &\multicolumn{1}{c}{P@0.8}  &\multicolumn{1}{c|}{P@0.9}  &\multicolumn{1}{c|}{0.5:0.95}      &\multicolumn{1}{c}{Overall}  &\multicolumn{1}{c}{Mean}   \\                 \midrule
        
	    \multicolumn{4}{c|}{Hu \etal \cite{hu2016segmentation}}     & \multicolumn{2}{c|}{2016}                  & \multicolumn{1}{c|}{34.8}  & \multicolumn{1}{c|}{23.6}  & \multicolumn{1}{c|}{13.3} & \multicolumn{1}{c|}{3.3 } & \multicolumn{1}{c|}{0.1} & 13.2 & \multicolumn{1}{c|}{47.4} & 35.0 \\
	    
		\multicolumn{4}{c|}{Li \etal \cite{li2017tracking}}               & \multicolumn{2}{c|}{2017}           & \multicolumn{1}{c|}{38.7}  & \multicolumn{1}{c|}{29.0}  & \multicolumn{1}{c|}{17.5} & \multicolumn{1}{c|}{6.6 } & \multicolumn{1}{c|}{0.1} & 16.3 & \multicolumn{1}{c|}{51.5} & 35.4 \\

		\multicolumn{4}{c|}{Gavrilyuk \etal \cite{gavrilyuk2018actor}}           & \multicolumn{2}{c|}{2018}       & \multicolumn{1}{c|}{47.5}  & \multicolumn{1}{c|}{34.7}  & \multicolumn{1}{c|}{21.1} & \multicolumn{1}{c|}{8.0 } & \multicolumn{1}{c|}{0.2} & 19.8 & \multicolumn{1}{c|}{53.6} & 42.1 \\

		\multicolumn{4}{c|}{Gavrilyuk \etal $\dagger$ \cite{gavrilyuk2018actor}}  & \multicolumn{2}{c|}{2018}      & \multicolumn{1}{c|}{50.0}  & \multicolumn{1}{c|}{37.6}  & \multicolumn{1}{c|}{23.1} & \multicolumn{1}{c|}{9.4 } & \multicolumn{1}{c|}{0.4} & 21.5 & \multicolumn{1}{c|}{55.1} & 42.6 \\

		\multicolumn{4}{c|}{ACGA \cite{wang2019asymmetric}}                        & \multicolumn{2}{c|}{2019}      & \multicolumn{1}{c|}{55.7}  & \multicolumn{1}{c|}{45.9}  & \multicolumn{1}{c|}{31.9} & \multicolumn{1}{c|}{16.0} & \multicolumn{1}{c|}{2.0} & 27.4 & \multicolumn{1}{c|}{60.1} & 49.0  \\

	    \multicolumn{4}{c|}{VT-Capsule \cite{mcintosh2020visual}}                  & \multicolumn{2}{c|}{2020}      & \multicolumn{1}{c|}{52.6}  & \multicolumn{1}{c|}{45.0}  & \multicolumn{1}{c|}{34.5} & \multicolumn{1}{c|}{20.7} & \multicolumn{1}{c|}{3.6} & 30.3 & \multicolumn{1}{c|}{56.8} & 46.0 \\

		\multicolumn{4}{c|}{CMDY \cite{wang2020context}}                        & \multicolumn{2}{c|}{2020}      & \multicolumn{1}{c|}{60.7}  & \multicolumn{1}{c|}{52.5}  & \multicolumn{1}{c|}{40.5} & \multicolumn{1}{c|}{23.5} & \multicolumn{1}{c|}{4.5} & 33.3 & \multicolumn{1}{c|}{62.3} & 53.1 \\

	    \multicolumn{4}{c|}{PRPE \cite{ning2020polar}}                        & \multicolumn{2}{c|}{2020}       & \multicolumn{1}{c|}{63.4}      & \multicolumn{1}{c|}{57.9}      & \multicolumn{1}{c|}{48.3}     & \multicolumn{1}{c|}{32.2}     & \multicolumn{1}{c|}{8.3}    & \multicolumn{1}{c|}{38.8} & \multicolumn{1}{c|}{66.1} & \multicolumn{1}{c}{52.9} \\

		\multicolumn{4}{c|}{CSTM \cite{hui2021collaborative}}                        & \multicolumn{2}{c|}{2021}      & \multicolumn{1}{c|}{65.4}  & \multicolumn{1}{c|}{58.9}  & \multicolumn{1}{c|}{49.7} & \multicolumn{1}{c|}{33.3} & \multicolumn{1}{c|}{9.1} & 39.9 & \multicolumn{1}{c|}{66.2} & \multicolumn{1}{c}{56.1} \\

	    \multicolumn{4}{c|}{M4TNVS \cite{zhao2022modeling} $\dagger$}                        & \multicolumn{2}{c|}{2022}       & \multicolumn{1}{c|}{64.5}      & \multicolumn{1}{c|}{59.7}      & \multicolumn{1}{c|}{52.3}     & \multicolumn{1}{c|}{37.5}     & \multicolumn{1}{c|}{13.0} & \multicolumn{1}{c|}{41.9} & \multicolumn{1}{c|}{67.3} & \multicolumn{1}{c}{55.8}  \\
	    
      \multicolumn{4}{c|}{ReferFormer \cite{wu2022language}}                     & \multicolumn{2}{c|}{2022}       & \multicolumn{1}{c|}{76.0}      & \multicolumn{1}{c|}{72.2}      & \multicolumn{1}{c|}{65.4}     & \multicolumn{1}{c|}{49.8}     & \multicolumn{1}{c|}{17.9} & \multicolumn{1}{c|}{48.6}  & \multicolumn{1}{c|}{72.3} & \multicolumn{1}{c}{48.6}  \\ 
       
         \multicolumn{4}{c|}{MTTR \cite{botach2022end}}                        & \multicolumn{2}{c|}{2022}       & \multicolumn{1}{c|}{75.4}      & \multicolumn{1}{c|}{71.2}      & \multicolumn{1}{c|}{63.8}     & \multicolumn{1}{c|}{48.5}     & \multicolumn{1}{c|}{16.9} & \multicolumn{1}{c|}{46.1}  & \multicolumn{1}{c|}{72.0} & \multicolumn{1}{c}{64.0}  \\

        \multicolumn{4}{c|}{LBDT \cite{ding2022language}}                      & \multicolumn{2}{c|}{2022}       & \multicolumn{1}{c|}{73.0}      & \multicolumn{1}{c|}{67.4}      & \multicolumn{1}{c|}{59.0}     & \multicolumn{1}{c|}{42.1}     & \multicolumn{1}{c|}{13.2} & \multicolumn{1}{c|}{47.2} & \multicolumn{1}{c|}{70.4} & \multicolumn{1}{c}{62.1}  \\  \midrule

	    \multicolumn{4}{c|}{Chen \etal \cite{chen2022weakly}*}                        & \multicolumn{2}{c|}{2022}       & \multicolumn{1}{c|}{33.3}      & \multicolumn{1}{c|}{26.3}      & \multicolumn{1}{c|}{17.6}     & \multicolumn{1}{c|}{7.9}     & \multicolumn{1}{c|}{0.6} & \multicolumn{1}{c|}{15.4} & \multicolumn{1}{c|}{31.7} & \multicolumn{1}{c}{30.7}  \\ 
	    
	    \multicolumn{4}{c|}{SimRVOS+Ours*}                        & \multicolumn{2}{c|}{2023}       & \multicolumn{1}{c|}{62.9}      & \multicolumn{1}{c|}{58.0}      & \multicolumn{1}{c|}{49.8}     & \multicolumn{1}{c|}{35.0}     & \multicolumn{1}{c|}{11.0} & \multicolumn{1}{c|}{40.0} & \multicolumn{1}{c|}{66.3} & \multicolumn{1}{c}{53.9}  \\ \bottomrule

	\end{tabular}

	\label{A2D}
\end{table*}

\begin{table*}[t!]
	\centering
	\scriptsize
	\renewcommand{\arraystretch}{1.0}
	\renewcommand{\tabcolsep}{4.3mm}
	\caption{Comparison with state-of-the-art methods on JHMDB-Sentences testing set. All methods adopt the best model trained on A2D-Sentences to directly eval on JHMDB-Sentences without finetuning.  $\dagger$ denotes adopting additional optical flow input. * denotes that the model is trained with weak annotations or without annotations.}
 
	\begin{tabular}{cccc|cc|ccccc|c|cc} 
	    \toprule 
        \multicolumn{4}{c|}{\multirow{2}{*}{Methods}}  & \multicolumn{2}{c|}{\multirow{2}{*}{Year}} & \multicolumn{5}{c|}{Precision}  & \multicolumn{1}{c|}{mAP}  & \multicolumn{2}{c}{IoU} \\ %\cline{7-14}
        
        \multicolumn{4}{c|}{}                                   &  \multicolumn{2}{c|}{}                                 &\multicolumn{1}{c}{P@0.5}               &\multicolumn{1}{c}{P@0.6} &\multicolumn{1}{c}{P@0.7}      &\multicolumn{1}{c}{P@0.8}  &\multicolumn{1}{c|}{P@0.9}  &\multicolumn{1}{c|}{0.5:0.95}      &\multicolumn{1}{c}{Overall}  &\multicolumn{1}{c}{Mean}   \\                 \midrule

	    \multicolumn{4}{c|}{Hu \etal \cite{hu2016segmentation}}     & \multicolumn{2}{c|}{2016}                & \multicolumn{1}{c|}{63.3} & \multicolumn{1}{c|}{35.0}  & \multicolumn{1}{c|}{8.5} & \multicolumn{1}{c|}{0.2} & \multicolumn{1}{c|}{0.0} &  \multicolumn{1}{c|}{17.8} & \multicolumn{1}{c|}{54.6} & \multicolumn{1}{c}{52.8} \\

		\multicolumn{4}{c|}{Li \etal \cite{li2017tracking}}             & \multicolumn{2}{c|}{2017}    & \multicolumn{1}{c|}{57.8}                & \multicolumn{1}{c|}{33.5} & \multicolumn{1}{c|}{10.3}  & \multicolumn{1}{c|}{0.6} & \multicolumn{1}{c|}{0.0} & \multicolumn{1}{c|}{17.3} &  \multicolumn{1}{c|}{52.9} & \multicolumn{1}{c}{49.1}  \\

		\multicolumn{4}{c|}{Gavrilyuk \etal \cite{gavrilyuk2018actor}}             & \multicolumn{2}{c|}{2018}     & \multicolumn{1}{c|}{69.9}                & \multicolumn{1}{c|}{46.0} & \multicolumn{1}{c|}{17.3}  & \multicolumn{1}{c|}{1.4} & \multicolumn{1}{c|}{0.0} & \multicolumn{1}{c|}{23.3} & \multicolumn{1}{c|}{54.1}  & \multicolumn{1}{c}{54.2}  \\

		\multicolumn{4}{c|}{ACGA \cite{wang2019asymmetric}}                       & \multicolumn{2}{c|}{2019}                & \multicolumn{1}{c|}{75.6} & \multicolumn{1}{c|}{56.4}  & \multicolumn{1}{c|}{28.7} & \multicolumn{1}{c|}{3.4} & \multicolumn{1}{c|}{0.0} &  \multicolumn{1}{c|}{28.9} & \multicolumn{1}{c|}{57.6} & \multicolumn{1}{c}{58.4} \\

	    \multicolumn{4}{c|}{VT-Capsule \cite{mcintosh2020visual}}                  & \multicolumn{2}{c|}{2020}                & \multicolumn{1}{c|}{67.7} & \multicolumn{1}{c|}{51.3}  & \multicolumn{1}{c|}{28.3} & \multicolumn{1}{c|}{5.1} & \multicolumn{1}{c|}{0.0} & \multicolumn{1}{c|}{26.1}  & \multicolumn{1}{c|}{53.5} & \multicolumn{1}{c}{55.0}  \\
		
		\multicolumn{4}{c|}{CMDY \cite{wang2020context}}                        & \multicolumn{2}{c|}{2020}                & \multicolumn{1}{c|}{74.2} & \multicolumn{1}{c|}{58.7}  & \multicolumn{1}{c|}{31.6} & \multicolumn{1}{c|}{4.7} & \multicolumn{1}{c|}{0.0}  & \multicolumn{1}{c|}{30.1} & \multicolumn{1}{c|}{55.4} & \multicolumn{1}{c}{57.6} \\
		
	    \multicolumn{4}{c|}{PRPE \cite{ning2020polar}}                        & \multicolumn{2}{c|}{2020}       & \multicolumn{1}{c|}{69.1}      & \multicolumn{1}{c|}{57.2}      & \multicolumn{1}{c|}{31.9}     & \multicolumn{1}{c|}{6.0}     & \multicolumn{1}{c|}{0.1}    & \multicolumn{1}{c|}{29.4} & \multicolumn{1}{c|}{-} & \multicolumn{1}{c}{-} \\

		\multicolumn{4}{c|}{CSTM \cite{hui2021collaborative}}                    & \multicolumn{2}{c|}{2021}               & \multicolumn{1}{c|}{78.3}                & \multicolumn{1}{c|}{63.9} & \multicolumn{1}{c|}{37.8}  & \multicolumn{1}{c|}{7.6} & \multicolumn{1}{c|}{0.0} & \multicolumn{1}{c|}{33.5}
		& \multicolumn{1}{c|}{59.8}  &  \multicolumn{1}{c}{60.4} \\
		
	    \multicolumn{4}{c|}{M4TNVS \cite{zhao2022modeling} $\dagger$}                        & \multicolumn{2}{c|}{2022}       & \multicolumn{1}{c|}{79.9}      & \multicolumn{1}{c|}{71.4}      & \multicolumn{1}{c|}{49.0}     & \multicolumn{1}{c|}{12.6}     & \multicolumn{1}{c|}{0.1} & \multicolumn{1}{c|}{38.6} & \multicolumn{1}{c|}{61.9} & \multicolumn{1}{c}{61.3}  \\ 
	         
        \multicolumn{4}{c|}{ReferFormer \cite{wu2022language}}                        & \multicolumn{2}{c|}{2022}       & \multicolumn{1}{c|}{93.3}      & \multicolumn{1}{c|}{84.2}      & \multicolumn{1}{c|}{61.4}     & \multicolumn{1}{c|}{16.4}     & \multicolumn{1}{c|}{0.3} & \multicolumn{1}{c|}{39.1} & \multicolumn{1}{c|}{70.0} & \multicolumn{1}{c}{69.3}  \\

      \multicolumn{4}{c|}{MTTR \cite{botach2022end}}                            & \multicolumn{2}{c|}{2022}       & \multicolumn{1}{c|}{93.9}      & \multicolumn{1}{c|}{84.2}      & \multicolumn{1}{c|}{61.4}     & \multicolumn{1}{c|}{16.4}     & \multicolumn{1}{c|}{0.3} & \multicolumn{1}{c|}{39.1} & \multicolumn{1}{c|}{70.0} & \multicolumn{1}{c}{69.3}  \\

        \multicolumn{4}{c|}{LBDT \cite{ding2022language}}                        & \multicolumn{2}{c|}{2022}       & \multicolumn{1}{c|}{86.4}      & \multicolumn{1}{c|}{74.4}      & \multicolumn{1}{c|}{53.3}     & \multicolumn{1}{c|}{13.2}     & \multicolumn{1}{c|}{0.0} & \multicolumn{1}{c|}{41.1} & \multicolumn{1}{c|}{64.5} & \multicolumn{1}{c}{65.8}  \\ \midrule

	    \multicolumn{4}{c|}{Chen \etal \cite{chen2022weakly}*}                        & \multicolumn{2}{c|}{2022}       & \multicolumn{1}{c|}{53.3}      & \multicolumn{1}{c|}{35.4}      & \multicolumn{1}{c|}{12.3}     & \multicolumn{1}{c|}{1.3}     & \multicolumn{1}{c|}{0.0} & \multicolumn{1}{c|}{17.6} & \multicolumn{1}{c|}{44.1} & \multicolumn{1}{c}{45.2}  \\ 
	    
	    \multicolumn{4}{c|}{SimRVOS+Ours*}                        & \multicolumn{2}{c|}{2023}       & \multicolumn{1}{c|}{82.0}      & \multicolumn{1}{c|}{67.3}      & \multicolumn{1}{c|}{41.4}     & \multicolumn{1}{c|}{8.9}     & \multicolumn{1}{c|}{0.1} & \multicolumn{1}{c|}{35.7} & \multicolumn{1}{c|}{63.2} & \multicolumn{1}{c}{62.7}  \\ \bottomrule

	\end{tabular}
	\label{JHMDB}
\end{table*}

\section{Future Research Directions}
In this Section, we propose some directions in RVOS that can be further explored in the future.
\begin{itemize}

    \item Building large-scale datasets for referring video object segmentation. As discussed in Section~\ref{sec:introduction}, RVOS methods are limited by the lack of sufficient data. Our efficient annotation scheme can reduce the cost of building a large-scale dataset and enables the possibility of creating large-scale RVOS datasets in the future.

    \item Exploring weaker annotation schemes. Although our annotation scheme is effective and achieves comparable performance to dense annotation, we acknowledge that there is still potential to optimize it. Future works can try to design methods that can learn from weaker annotation to further reduce the burden of annotation.

    \item Introducing pretrained language-vision models into RVOS. Language-vision pretraining \cite{radford2021learning} achieves remarkable performance on many tasks. Exploring how to leverage these pretrained language-vision models may further improve performance and largely reduce the reliance on densely-annotated RVOS datasets.

\end{itemize}% √ 

\section{Conclusion}
In the paper, we investigate different paradigms of weak supervision for RVOS and propose to only label the frame where the target object first appears with a mask and other frames with bounding boxes, which reduces the cost of annotation by leaps and bounds. Based on this scheme, we propose a language-guided cross frame segmentation method to fully take advantage of both mask and bounding box annotations. Then, a bi-level contrastive learning method is designed to promote the learning of discriminative feature representation at the pixel level. Experiments show that our model achieves competitive performance and even outperforms fully-supervised RVOS methods. 

% Finally, we admit that we do not construct a larger RVOS dataset based on our weak annotation scheme, which can be left to future works.

\ifCLASSOPTIONcaptionsoff
  \newpage
\fi

\bibliographystyle{IEEEtran}
\bibliography{refer}

% \vspace{-5.0em}
\begin{IEEEbiography}[{\includegraphics[width=1in,height=1.25in,clip,keepaspectratio]{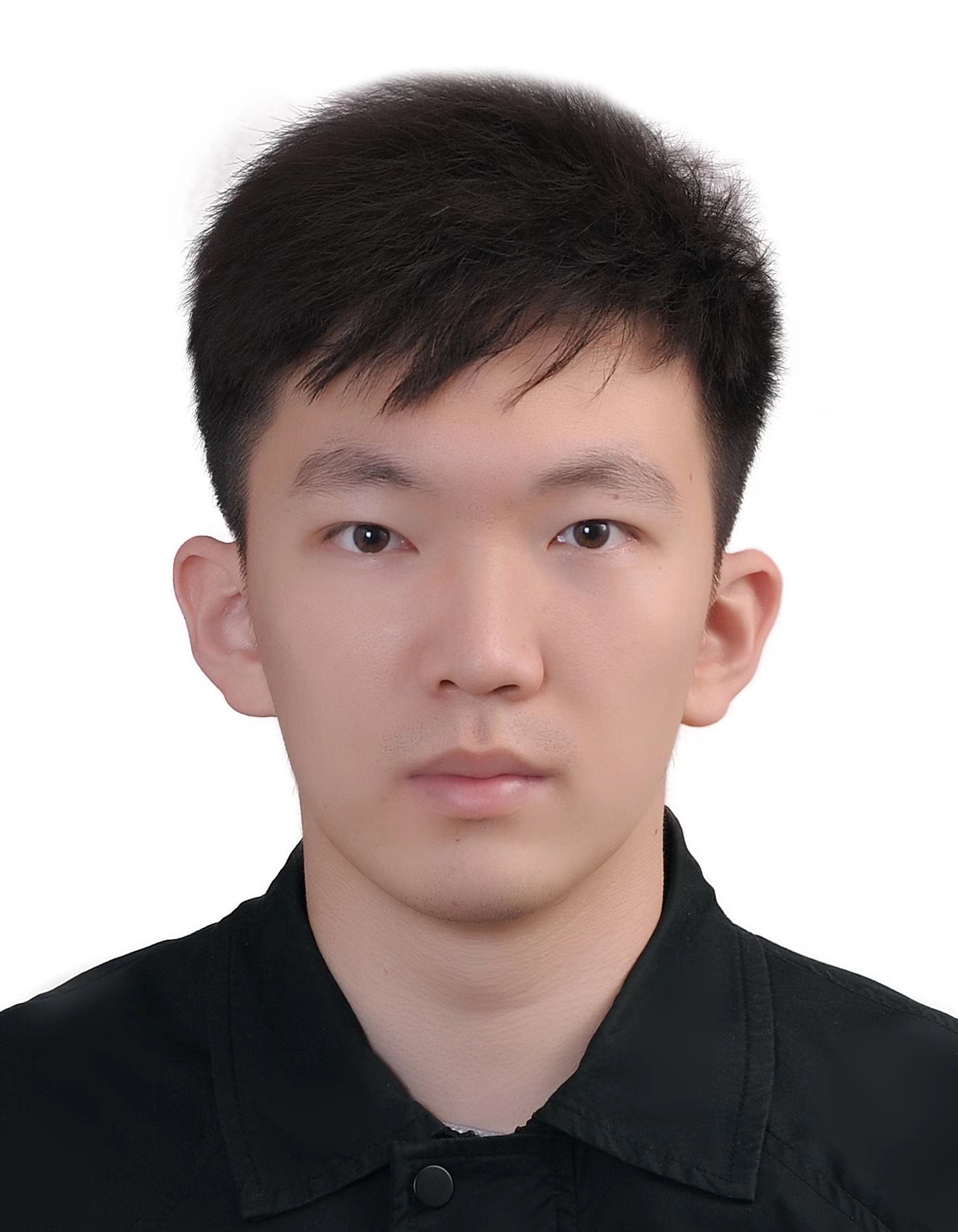}}]
{Wangbo Zhao} currently is a Ph.D. student at the National University of  Singapore. He received his Bachelor and Master's degrees from Northwestern Polytechnical University, Xi’an, China in 2019 and 2022, respectively. His research interest is computer vision, especially in video object segmentation and multi-modal model. 
\end{IEEEbiography}

% \vspace{-2.0em}
\begin{IEEEbiography}[{\includegraphics[width=1in,height=1.25in,clip,keepaspectratio]{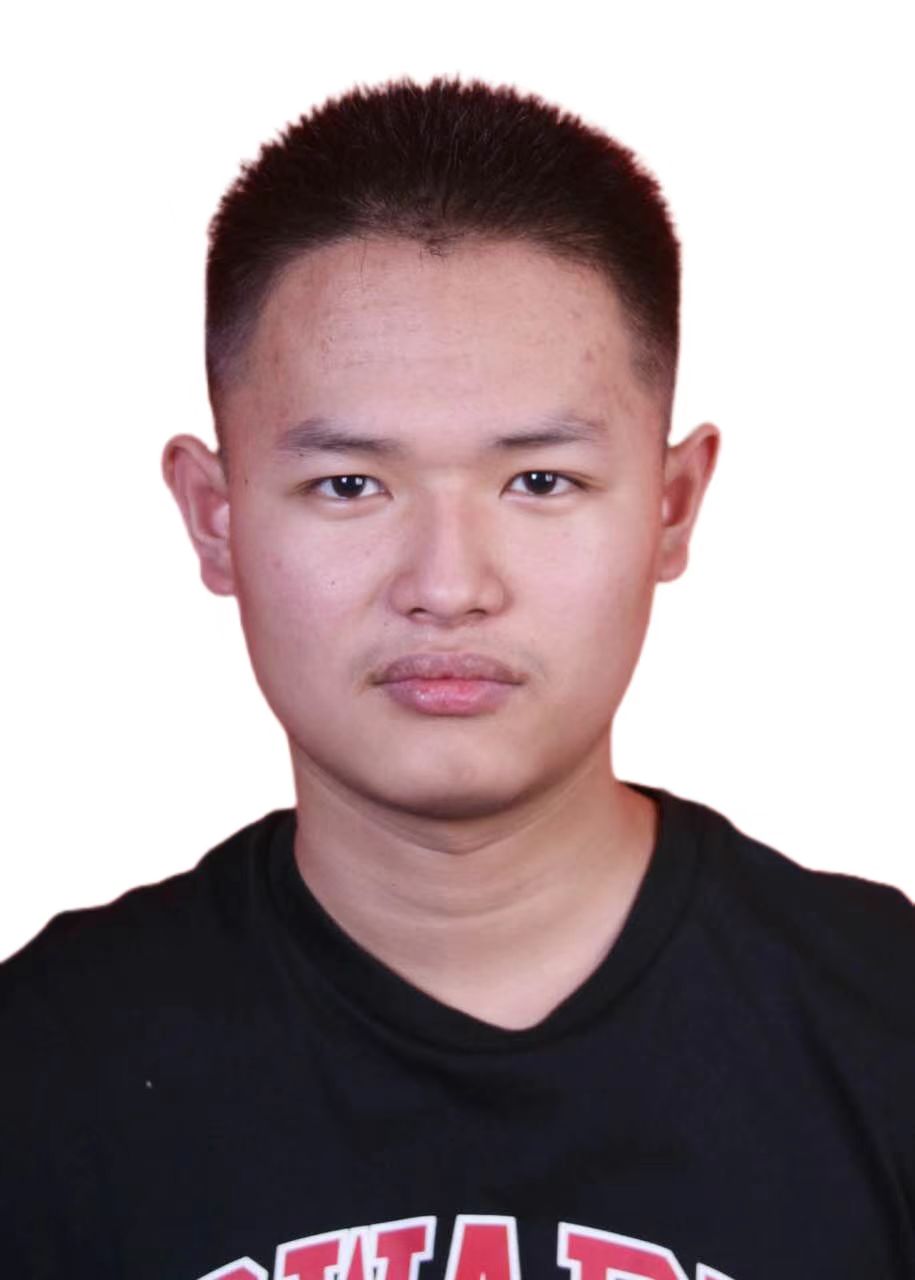}}]
{Kepan Nan} received the bachelor’s degree from Northwestern Polytechnical University, Xi’an, China, in 2017, where he is currently pursuing the
master’s degree. His research interests include computer vision and deep learning, especially on few-shot learning and video object segmentation.
\end{IEEEbiography}

% \vspace{-5.0em}
\begin{IEEEbiography}[{\includegraphics[width=1in,height=1.25in,clip,keepaspectratio]{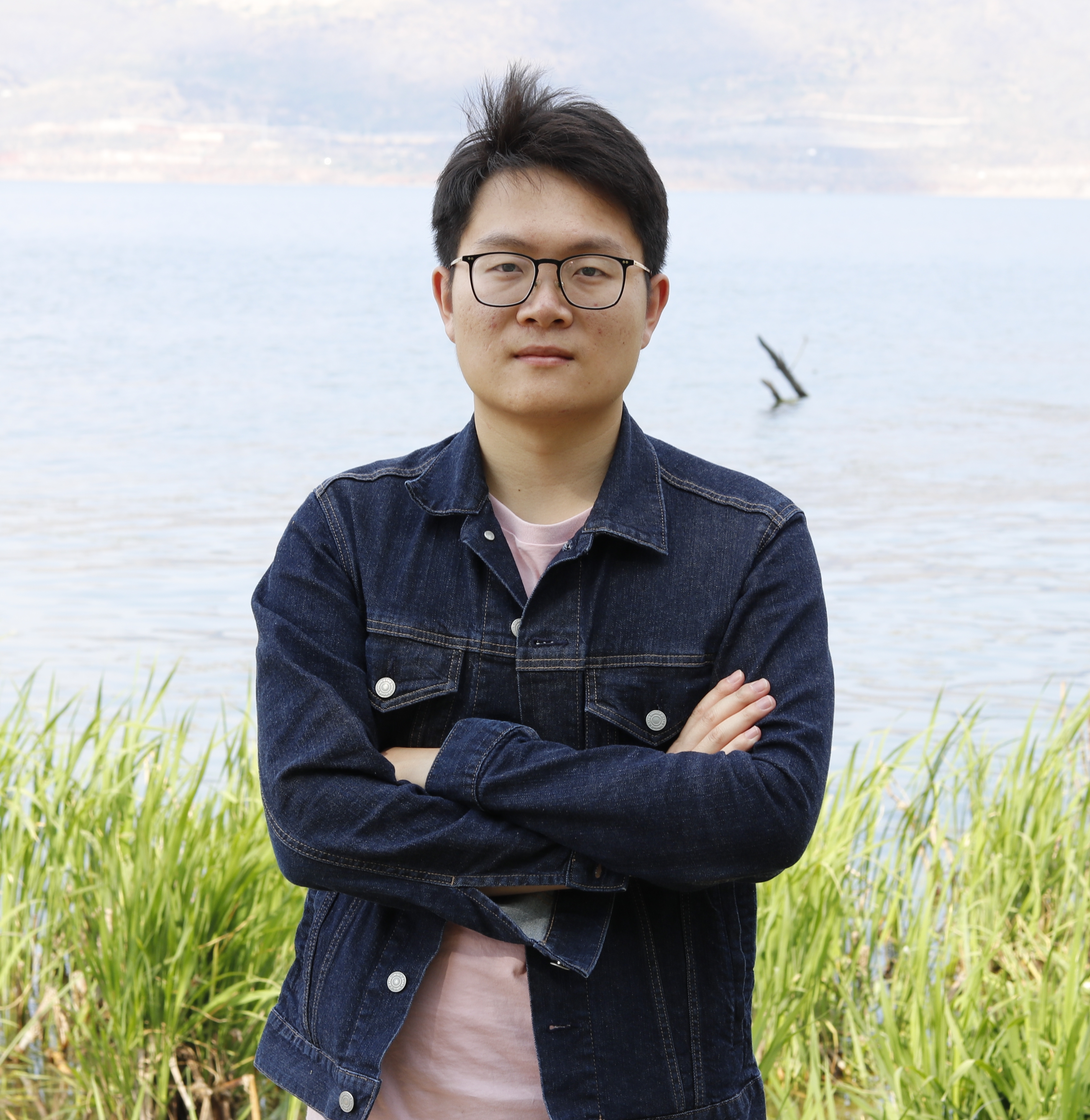}}]
{Songyang Zhang} received the Ph.D. degree in Computer Science at the University of Chinese Academy of Science, in the joint program at PLUS Lab, ShanghaiTech University, supervised Prof. Xuming He in 2022. He is currently work as Postdoctoral Fellow at Shanghai AI Laboratory, Shanghai, China. His research interests concern computer vision and deep learning.
\end{IEEEbiography}

% \vspace{-3.0em}
\begin{IEEEbiography}[{\includegraphics[width=1in,height=1.25in,clip,keepaspectratio]{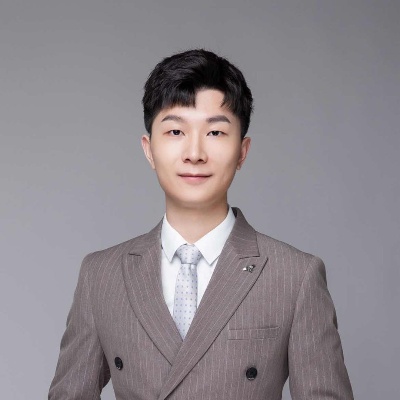}}]
{Kai Chen} is currently a director at SenseTime, and also a Research Scientist and PI at Shanghai AI Laboratory. He is leading the OpenMMLab team for open-source ecosystem and a R-D team for business service. The teams target at developing state-of-the-art computer vision algorithms for research and industrial applications, as well as building influential open-source platforms. He received the PhD degree in The Chinese University of Hong Kong in 2019, under the supervision of Prof. Dahua Lin and Chen Change Loy at MMLab. Before that, he received the B.Eng. degree from Tsinghua University in 2015.
\end{IEEEbiography}

% \vspace{-10.0em}
\begin{IEEEbiography}[{\includegraphics[width=1in,height=1.5in,clip,keepaspectratio]{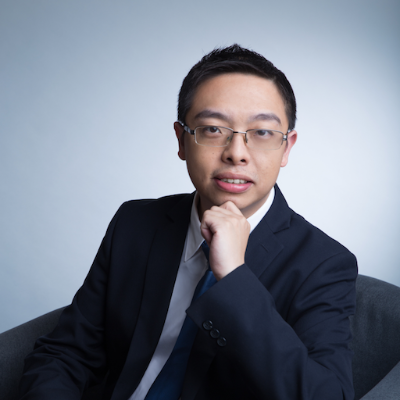}}]
{Dahua Lin} received the B.Eng. degree from the University of Science and Technology of China, Hefei,
China, in 2004, the M. Phil. degree from The Chinese University of Hong Kong, Hong Kong, in 2006, and
the Ph.D. degree from the Massachusetts Institute of Technology, Cambridge, MA, USA, in 2012. From
2012 to 2014, he was a Research Assistant Professor with Toyota Technological Institute at Chicago,
Chicago, IL, USA. He is currently an Associate Professor with the Department of Information Engineering, The Chinese University of Hong Kong (CUHK), and the Director of CUHK-SenseTime Joint Laboratory. His research interests include computer vision and machine learning. He serves on the Editorial Board of the International Journal of Computer Vision. He is also the Area Chair of the
multiple conferences, including ECCV 2018, ACM Multimedia 2018, BMVC 2018, CVPR 2019, BMVC 2019, AAAI 2020, CVPR 2021, and CVPR 2022.
\end{IEEEbiography}

\vspace{-3.0em}
\begin{IEEEbiography}[{\includegraphics[width=1in,height=1.5in,clip,keepaspectratio]{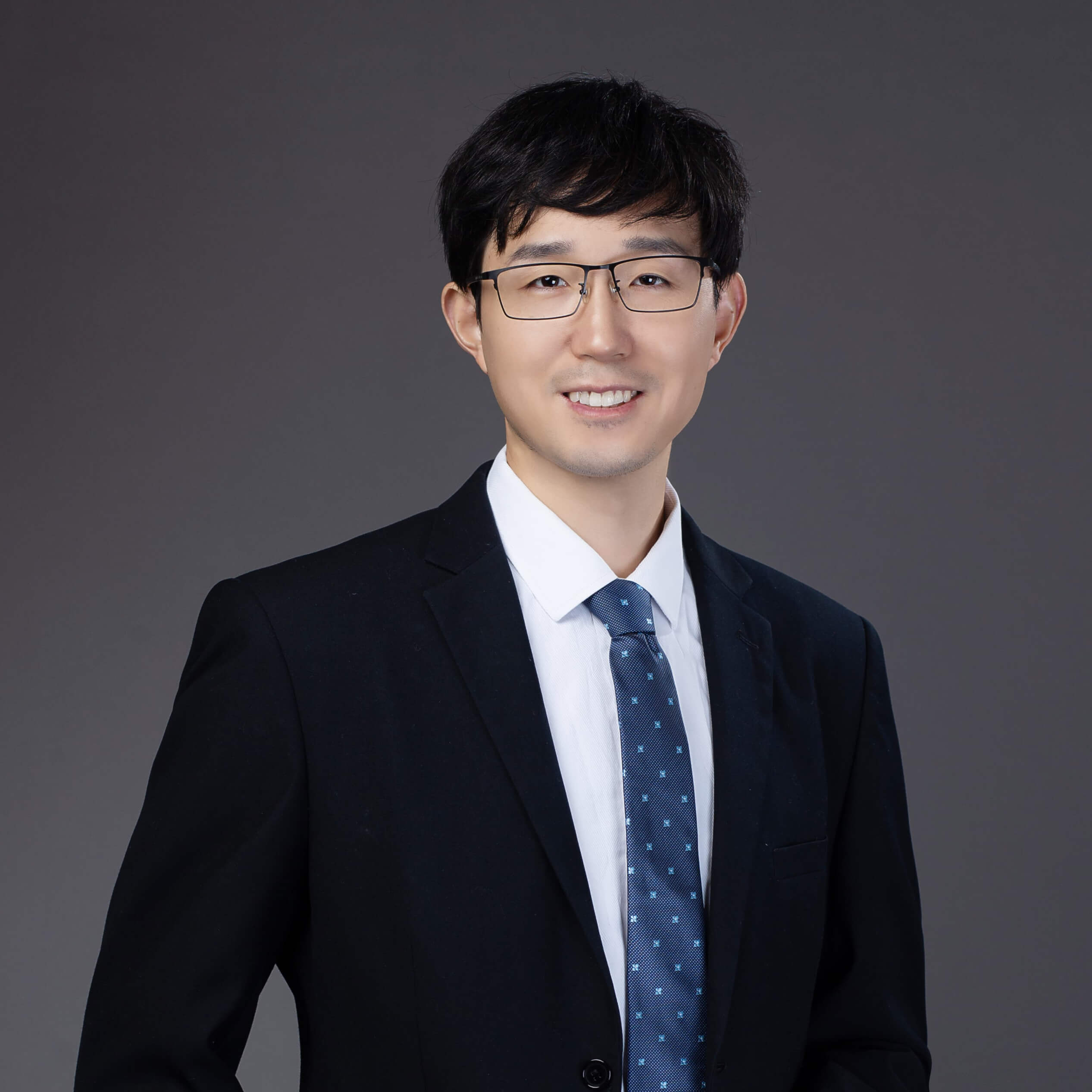}}]
{Yang You} received the PhD degree in computer science from UC Berkeley. He is a presidential
young professor with the National University of Singapore. His research interests include parallel/distributed algorithms, HPC, and ML. He is a winner of IPDPS 2015 and ICPP 2018 Best Paper Award. He is the founder and chairman of HPCAI Technology.
\end{IEEEbiography}

\end{document}